\pdfoutput=1
\documentclass[11pt]{article}
\usepackage[final]{acl}
\usepackage{tabularx}
\usepackage{booktabs}
\usepackage{siunitx} 
\usepackage{array}
\usepackage{makecell}
\usepackage{threeparttable}
\usepackage{adjustbox}
\usepackage{multicol}
\usepackage{refcount}
\usepackage{subcaption}
\usepackage{placeins}
\usepackage{hyperref}

\usepackage{xurl}
\usepackage{times}
\usepackage{latexsym}
\usepackage[T1]{fontenc}
\usepackage[utf8]{inputenc}
\usepackage{microtype}
\usepackage{inconsolata}
\usepackage{kotex}
\usepackage{threeparttablex}

\usepackage{amsmath}

\usepackage{url}
\usepackage{graphicx}

\usepackage[most]{tcolorbox}
\tcbset{
  promptstyle/.style={
    colback=gray!3, colframe=black!15,
    boxrule=0.4pt, left=6pt,right=6pt,top=6pt,bottom=6pt
  }
}

\newcommand{\ours}{\textsc{PLAT}}
\newcommand{\Ours}{\textsc{Predicting the Legitimacy of punitive Additional Tax}}
\newcommand{\oursthree}{\textsc{PLAT-mc}}
\newcommand{\ourstwo}{\textsc{PLAT-mc$_2$}}

\newcommand{\oursfour}{\textsc{PLAT-mc$_R$}}
\newcommand{\oursessay}{\textsc{PLAT-e}}

\title{Taxation Perspectives from Large Language Models:\\A Case Study on Additional Tax Penalties}
\author{
Eunkyung Choi \\ \texttt{rmarud202@uos.ac.kr} \And
Youngjin Suh \\ \texttt{yjsuh624@uos.ac.kr} \And 
Siun Lee \\ \texttt{nojeo01@uos.ac.kr} \And
Hongseok Oh$^{\dagger}$ \\ \texttt{cxv0519@uos.ac.kr} \AND
Juheon Kang \\ \texttt{2715wngjs@uos.ac.kr} \And
Won Hur \\ \texttt{wonh99@uos.ac.kr} \And
Hun Park$^{*}$ \\ \texttt{phn@uos.ac.kr} \And
Wonseok Hwang$^{\dagger,}$\thanks{Corresponding authors. $\dagger$: Part of the work was done while the authors were at LBOX.}\\ \texttt{wonseok.hwang@uos.ac.kr}
\AND University of Seoul
}

\begin{document}
\maketitle
\begin{abstract}
How capable are large language models (LLMs) in the domain of taxation?
Although numerous studies have explored the legal domain, 
research dedicated to taxation remains scarce. Moreover, the datasets used in these studies are either simplified, failing to reflect the real-world complexities, or not released as open-source.
To address this gap, we introduce \ours, a new benchmark designed to assess the ability of LLMs to predict the legitimacy of additional tax penalties.
\ours\ comprises 300 examples: (1) 100 binary-choice questions, (2) 100 multiple-choice questions, and (3) 100 essay-type questions, all derived from 100 Korean court precedents.
\ours\ is constructed to evaluate not only LLMs' understanding of tax law but also their performance in legal cases that require complex reasoning beyond straightforward application of statutes.
Our systematic experiments with multiple LLMs reveal that (1) their baseline capabilities are limited, especially in cases involving conflicting issues that require a comprehensive understanding (not only of the statutes but also of the taxpayer's circumstances), and (2) LLMs struggle particularly with the “AC” stages of “IRAC”\footnote{IRAC: Issue, Rule, Application, Conclusion. A structured method of legal analysis used to organize legal arguments and reasoning.} even for advanced reasoning models like o3, which actively employ inference-time scaling. The dataset is publicly available at \href{https://huggingface.co/collections/sma1-rmarud/plat-predicting-the-legitimacy-of-punitive-additional-tax}{Hugging Face}.
\end{abstract}

\section{Introduction}
Large Language Models (LLMs) have demonstrated promising results across various domains. 
Among them, the legal domain was an early focus since OpenAI's demonstration that GPT-4 passes the U.S. Uniform Bar Exam \citep{martinez2023reeval_bar_exam_gpt4}. 
To more robustly assess LLMs' capabilities in the legal domain beyond bar exam, where questions may follow certain patterns, many studies have proposed benchmarks \citep{guha2023legalbench,fei-etal-2024-lawbench,kimyeeun2024femnlp-kbl} and analyzed LLM performance \citep{magesh2024lhallucinationfree,kang2023-chatgpt-irac,trautmann-etal-2024nllp-measuring-groundedness,chalkidis-2023-lexglue-chatgpt}.

However, in the taxation domain, despite its close relationship with the legal field, there has been little research on assessing LLM capabilities.
Previous studies have primarily focused on relatively simple questions that can be answered mostly based on deductive application of statutes \citep{holzenberger2020SARAdataset,Nay2024LLM-as-tax-attorney}, or have used real-world datasets without releasing them as open source, making reproduction difficult \citep{harvey2024tax-blog,Zhong2024ieeeaccess_taxqa_system_gov}.
Also, with rapid progress in LLMs and advances in LLM-based agents (or test-time scaling) \citep{openai2024o1,guo2025deepseek-r1}, issues such as deductive reasoning~\cite{lee2025naccl_symba} or simple calculation errors can now be easily mitigated using external tools. This suggests that more advanced benchmarks may be necessary for comprehensive evaluation in the taxation domain.

Here, we introduce \ours\footnote{\Ours}, a benchmark consisting of 300 questions derived from Korean precedents concerning the legitimacy of additional tax penalties. Article 48 of the Korean Framework Act on National Taxes\footnote{ \url{https://elaw.klri.re.kr/kor_service/lawTwoView.do?hseq=28738}} allows exemptions from penalty taxes in cases of {\it justifiable reasons}, but the statute does not explicitly define what constitutes such reasons. In practice, this gap has been filled by judicial precedents, where the courts have elaborated on and recognized specific circumstances as qualifying “justifiable reasons”. Thus, we designed \ours\ to assess LLMs' tax law comprehension and inferential ability--particularly in scenarios where issues cannot be resolved by statute alone and call for real-world legal judgment, including leniency, compassion, and discretionary reasoning typically exercised by human judges--such as weighing competing legal principles or determining whether it is reasonable to expect a taxpayer to recognize and comply with the law.

Our experiments with two open-source LLMs--Qwen3~\citep{yang2025qwen3technicalreport},
Exaone3.5~\citep{research2024exaone35serieslarge}--alongside eight commercial LLMs (GPT 4o, 4.1, o1, o3, o3-mini, Claude 3.7, Gemini2.5-flash-preview and 2.5-pro-preview) show that the strongest reasoning models, o3 and Gemini2.5-pro, achieve an $F_1$ score of 79\% on \ours. 
A detailed analysis reveals that, while LLMs perform well on relatively simple problems, their accuracy declines when a more inferential understanding is required.\footnote{For instance, all LLMs correctly recognize that ignorance or misunderstanding by taxpayers cannot serve as a justifiable reason. However, when the misunderstanding originates from an incorrect statement of opinion by the tax authority, accuracy drops due to a conflict between two legal principles: (1) the final responsibility lies with the taxpayer, vs (2) the principle of protection of legitimate expectations.}

To address this issue, we adopt the IRAC framework and investigate how LLM performance varies under the following conditions: (1) enabling retrieval-augmented generation (RAG), (2) providing the “Application” and “Conclusion” stages, and (3) introducing more complex essay-type questions to explore all the “IRAC” factors.


Our findings reveal that
(1) LLMs are relatively proficient at identifying the “Issue”; 
(2) consistent with prior work, they struggle to identify the correct “Rule” due to hallucinations \cite{dahl2024largelegalfiction}, though this can be mitigated with agentic RAG;
(3) LLMs underperform in the “Application” and “Conclusion” stages. Without inference-time scaling, they often hesitate to proceed, resulting in low recall, whereas with inference-time scaling, they do continue but frequently reach incorrect conclusions;
(4) when the “Answer” (Conclusion) and a corresponding simplified “Reason” (Application) are provided as a starting point, LLM accuracy improves significantly, highlighting the potential of backward-chaining reasoning in legal contexts;
(5) regardless of inference-time techniques or task formats, final “Conclusion” accuracy remains limited. 


In summary, our contributions are 
\begin{itemize}
    \item We propose a new dataset, \ours, to evaluate LLMs' understanding of tax law, particularly in legal cases that cannot be resolved solely by referencing statutes.
    \item We assess ten LLMs and find that, while they exhibit some competence, their performance is limited, especially in comprehending legal cases at the “Conclusion” stage even with inference-time scaling.
\end{itemize}
Our datasets, both the original Korean and English translated version, will be released to the community under a CC BY-NC license.

\section{Related Work}

\subsection{NLP in the Taxation domain}

\citet{Nay2024LLM-as-tax-attorney} study GPT-4's capability in handling tax law inquiries with and without retrieval augmented generation (RAG). Their study uses synthetically generated multiple-choice questions based on templates, where answers can be derived from either the Treasury Regulations under the U.S. Code of Federal Regulations (CFR) or Title 26 of the U.S. Code. The dataset has not been released.

\citet{holzenberger2020SARAdataset} develop SARA, a statutory reasoning dataset constructed from a simplified version of the U.S. Internal Revenue Code. The dataset consists of two tasks: determining entailment relations and calculating tax amounts based on given statutes and cases. Since all questions can be answered mostly through deductive reasoning from the given statutes, the dataset primarily comprises relatively simple questions.

\citet{Zhong2024ieeeaccess_taxqa_system_gov} suggest a retrieval-based 
 LLM system designed to answer tax-related questions typically handled by tax departments. The dataset has not been released. 
 
Compared to these studies, our dataset consists of 300 manually constructed examples, supervised by tax professionals.\footnote{Authors of this paper. Two professors, one PhD student in Taxation Science and one undergraduate student holding a national tax license.}
\ours\ is particularly distinct from previous datasets in that its questions cannot be answered solely by referencing statutes. Instead, they require an understanding of tax law and complex reasoning about real-world situations.

\subsection{Agent}
LLM-based AI agents are being rapidly developed. Unlike vanilla LLMs, which simply generate output text based on input text, LLM-based agents can enhance their capabilities by leveraging external tools for knowledge retrieval (e.g., search engine), improving reasoning (e.g., logic solver \citep{lee2025naccl_symba}), or refining internal knowledge through memory and self-reasoning processes. These processes can be iteratively orchestrated by the LLMs themselves. Below, we highlight a few representative works.

\citet{yao2023neurips_tot} introduce the Tree-of-Thoughts inference algorithm, which allows LLMs to generate and navigate multiple reasoning paths unlike Chain-of-Thought \citep{wei2022neurips_cot}, which follows only a single path. 

\citet{yao2023react} propose ReAct, which integrates reasoning and planning (such as action generation and document retrieval). The inference process is formalized into three key steps: thought (planning), action (tool calling), and observation (interpreting tool-generated results).


\citet{roucher2025smolagents} introduce \texttt{smolagents}, open-source framework designed for simplicity and seamless Python code integration. The smolagents framework is employed in this study.

\section{Datasets}
\subsection{Motivation}
In Korea, an additional penalty tax can be applied to all 25 types of taxes. It is an extra economic burden imposed on taxpayers who fail to properly file or pay their taxes, in addition to the original tax liability.\footnote{Korea’s additional tax penalties function as punitive measures, often imposing heavy burdens even for minor taxpayer errors. Their growing scale (e.g., in 2024, VAT (Value-Added-Tax) penalties rose by 40\% ($\sim$\$400m)) underscores the need to evaluate their legitimacy, a task for which \ours\space aims to ``plat'' (balance) state authority and taxpayer burden onto the same analytical plane, thereby strengthening the protection of taxpayers’ rights.} 
However, when objective circumstances prevent a taxpayer from fulfilling their obligations, it is more appropriate not to impose the penalty tax, even if a legal basis exists.

Indeed, section 2 of Article 48 of Korean Framework Act on National Taxes explicitly states that a penalty tax shall not be imposed if there is a “justifiable reason”. However, this phrase is an indeterminate concept lacking a clear scope, requiring interpretation in specific cases \citep{kim2008kor_broad_vague_concept_in_tax_law,yang2024kor_additional_tax,Park2019kor_borrowing_concept_tax_law}.
In such situations, precedents act as a interpretative standards.\footnote{Although Korean legal system is rooted in civil law system, higher courts’ decisions, especially those of the Supreme Court, are typically followed by lower courts.}

Thus, resolving whether a penalty tax is justified requires not only referencing the statutes but also comprehensively understanding the individual situation like human judges.\footnote{
This also indicate that the topic studied here reflects a real-world legal decision-making, an area in which current LLMs show limited capability. Accordingly, our work can serve as a benchmark not only for tax-related statutory reasoning but also for evaluating how effectively LLMs apply tax law beyond the statutes by considering individual factual circumstances.
}
In this regard, we build \ours, a benchmark constructed from 100 Korean precedents--50 justifiable, 50 not justifiable cases--addressing the legitimacy of additional tax penalties.



\subsection[Dataset Construction]{Dataset Construction}
\label{sec:dataset-construction}

We first collect relevant precedents using the commercial Korean legal search engine LBOX\footnote{\url{https://lbox.kr/v2}}, searching with the keyword “additional penalty tax”. The query returned approximately 20k precedents. To further refine the dataset, we added the keyword “justifiable reasons”, reducing the target cases to 3.7k. Finally, we excluded cases containing the keyword  “gift tax”, as such cases primarily focus on the issue related to the method of tax calculation. This results in total 2.8k candidate pool. 

From the candidate pool, we sampled 10 precedents and extracted facts and claims from them using o3. Then a tax professional (one of our authors) manually evaluated based on the following criteria:
\begin{itemize}
    \item Well-defined task: Does the input contain sufficient information to answer the question? Are the main issues of the selected cases related to an additional penalty tax?
    \item Information leakage: Is there any unintended disclosure of the court decision in the input?
    \item Hallucination: Are there any inaccuracies or fabricated information in the extracted facts and claims?
    \item Legal Correctness: Are the labels extracted from court rulings consistent with the actual court decisions?
\end{itemize}

Under the result of evaluation, we curated final 100 examples, with an equal split: 50 cases where the court ruled the exemption from penalty tax was justified, and 50 cases where the court decided that the exemption was not. Each example required approximately 30--40 minutes for evaluation, resulting in a total of  50--67 hours of expert review time.
Based on this, we built two multiple-choice (\oursthree, \oursfour) and one essay-type (\oursessay) QA datasets. (\ours\ examples are in Appendix \ref{platex}.)
\begin{itemize}
    \item \oursthree: Each question provides two answer choices--“lawful” and “unlawful”. In practice, especially in the legal domain where reliability is critical, it is important for models to be able to express uncertainty. Therefore, we included an additional “don’t know” option for cases where the model is uncertain. (Table \ref{tbl_data_example_eng} in Appendix \ref{emc})
    \item \oursfour: We labeled choices according to court's logic and judge's decision in precedents. Each option includes not only whether the judgment is lawful or unlawful, but also the key rationale behind the judge’s decision. All 400 options were manually labeled, evaluated, and double-checked by two tax professionals\footnote{All annotations for the dataset were performed by an undergraduate student holding a national tax license and by a PhD student in Taxation Science (Authors of this paper). The inter-annotator agreement rate was 92\%.}. (Table \ref{tbl_4_choices} in Appendix \ref{emc4})
    \item \oursessay: Each essay question follows the format of the second-round essay-style exam for the Korean Certified Tax Accountant (CTA). We considered the court’s reasoning and the judge’s final decision as the reference answer. The task-specific rubrics are designed based on the IRAC framework, curated through discussion among three tax professionals who also served as human judges for the LLM essay responses. (Appendix \ref{ee})
\end{itemize}

Additional details about \ours\ construction and the annotation GUI are provided in Appendix~\ref{datasetannotaion}.

\newcommand{\stddev}[1]{\textcolor{black!60}{\scriptstyle (#1)}}

\begin{table*}[th!]
  \caption{F1 scores on \oursthree. Numbers in parenthesis indicate standard deviation from three independent runs. }
  \label{tab:base_model_f1_plat_3mc}
  \centering
  \small 
  \begin{threeparttable}
    \setlength{\tabcolsep}{3pt}%
    \renewcommand{\arraystretch}{0.95}%

    \begin{adjustbox}{max width=\textwidth}
    \begin{tabular}{@{}p{2.3cm}|c|c|c|c|c|c|c|c|c@{}}
      \toprule
      \textbf{Model} & \textbf{F1} & \textbf{P} & \textbf{R} & \textbf{F1-easy} & \textbf{P-easy} & \textbf{R-easy} & \textbf{F1-hard} & \textbf{P-hard} & \textbf{R-hard}\\
      \toprule
      Exaone3.5-32B &0.55 &0.69 &0.46 &0.72 &0.86 &0.61 &0.31 &0.20 &0.62\\
      Qwen3-32B &0.75 &0.60 &0.98 &0.80 &0.67 &0.95 &0.67 &0.50 &1.00\\
      \midrule
      Claude3.7-sonnet &0.74 &0.63 &0.91 &0.68 &0.53 &0.94 &0.75 &0.65 &0.89\\
      Gemini2.5-flash &0.73 &0.63 &0.88 &0.78 &0.69 &0.89 &0.70 &0.59 &0.88\\
      GPT-4o &0.70 &0.62 &0.81 &0.88 &0.83 &0.92 &0.45 &0.32 &0.80\\
      GPT-4.1 &0.65 &0.66 &0.64 &0.75 &0.76 &0.74 &0.49 &0.42 &0.64\\
      \midrule
      Qwen3-32B$^{\dagger)}$ & $0.72_{\stddev{\pm0.05}}$ & $0.57_{\stddev{\pm0.04}}$ & $0.96_{\stddev{\pm0.06}}$
      & $0.60_{\stddev{\pm0.02}}$ & $0.44_{\stddev{\pm0.02}}$ & $0.94_{\stddev{\pm0.06}}$
      & $0.69_{\stddev{\pm0.03}}$ & $0.53_{\stddev{\pm0.03}}$ & $0.97_{\stddev{\pm0.03}}$ \\
      o3-mini & $0.69_{\stddev{\pm0.01}}$ & $0.53_{\stddev{\pm0.01}}$ & $0.97_{\stddev{\pm0.03}}$ & $0.90_{\stddev{\pm0.03}}$ & $0.85_{\stddev{\pm0.04}}$ & $0.95_{\stddev{\pm0.01}}$
      & $0.46_{\stddev{\pm0.03}}$ & $0.31_{\stddev{\pm0.02}}$ & $0.95_{\stddev{\pm0.02}}$ \\
      o1 & $0.75_{\stddev{\pm0.04}}$ & $0.62_{\stddev{\pm0.04}}$ & $0.96_{\stddev{\pm0.02}}$ & $0.92_{\stddev{\pm0.01}}$ & $0.86_{\stddev{\pm0.02}}$ & $0.99_{\stddev{\pm0.01}}$
      & $0.62_{\stddev{\pm0.07}}$ & $0.47_{\stddev{\pm0.07}}$ & $0.94_{\stddev{\pm0.02}}$ \\
      \textbf{o3} & $\textbf{0.79}_{\stddev{\pm0.03}}$ & $0.65_{\stddev{\pm0.04}}$ & $1.00_{\stddev{\pm0.00}}$ & $0.83_{\stddev{\pm0.02}}$ & $0.71_{\stddev{\pm0.03}}$ & $1.00_{\stddev{\pm0.00}}$ & $0.77_{\stddev{\pm0.05}}$ & $0.62_{\stddev{\pm0.08}}$ & $1.00_{\stddev{\pm0.00}}$ \\
      Gemini2.5-flash$^{\dagger\text{i}}$ & $0.74_{\stddev{\pm0.02}}$ & $0.62_{\stddev{\pm0.03}}$ & $0.91_{\stddev{\pm0.01}}$ & $0.78_{\stddev{\pm0.03}}$ & $0.69_{\stddev{\pm0.05}}$ & $0.90_{\stddev{\pm0.02}}$ & $0.71_{\stddev{\pm0.04}}$ & $0.59_{\stddev{\pm0.05}}$ & $0.91_{\stddev{\pm0.02}}$ \\
      \textbf{Gemini2.5-pro}$^{\dagger\text{i}}$ & $\textbf{0.79}_{\stddev{\pm0.01}}$ & $0.68_{\stddev{\pm0.01}}$ & $0.95_{\stddev{\pm0.01}}$ & $0.78_{\stddev{\pm0.05}}$ & $0.65_{\stddev{\pm0.06}}$ & $0.96_{\stddev{\pm0.02}}$ & $0.80_{\stddev{\pm0.03}}$ & $0.70_{\stddev{\pm0.04}}$ & $0.95_{\stddev{\pm0.01}}$\\
      \bottomrule
    \end{tabular}
    \end{adjustbox}
    {\setlength{\emergencystretch}{2pt}%
    \begin{tablenotes}[para,flushleft]\footnotesize
    \item[$\dagger$] Reasoning mode, $^{\text{i}}$ thinking-budget: 4096
    \end{tablenotes}}
    
  \end{threeparttable}
\end{table*}

\section{Experiments}
\subsection{Models}
We used two open-weight LLMs (Qwen3-32B\footnote{Qwen3-32B}, LG EXAONE3.5-32B\footnote{EXAONE-3.5-32B-Instruct}) and eight commercial LLMs (GPT 4o, 4.1, o1, o3, o3-mini\footnote{gpt-4o-2024-11-20, gpt-4.1-2025-04-14, o1-2024-12-17, o3-2025-04-16, o3-mini-2025-01-31}, Claude 3.7 sonnet\footnote{claude-3-7-sonnet-20250219}, Gemini2.5-flash-preview and 2.5-pro-preview\footnote{gemini-2.5-flash-preview-05-20, gemini-2.5-pro-preview-06-05. We evaluated \oursessay\ using the stable (non-preview) version, as the preview version is no longer available.}).
For all experiments with non-reasoning models, we set the temperature to 0. For reasoning models that do not support temperature settings, we conducted three independent runs.
\subsection{Retriever Setup}
\label{subsec:retrieval}
We used Pyserini~\citep{Lin_etal_SIGIR2021_Pyserini} with the BM25 algorithm with default hyperparameters. We limited the retriever to the top-3 documents, based on initial experiments comparing top-1, 3, 5, and 10. The retrieval pool comprises 100 precedents related to additional tax penalties and 4,042 articles related to Korean tax law.
Articles are filtered from the Korean Statutes Corpus \citep{kimyeeun2024femnlp-kbl}. The precedents in the retrieval pool were processed using o3 to extract facts and claims. The source precedent for each question was excluded from the retrieval pool to prevent information leakage.
\texttt{smolagents} was employed \citep{roucher2025smolagents} to build LLM agents.

\subsection{Metrics}
In \oursthree\ and \oursfour, a model first generates (selects) an answer among possible choices followed by accompanying rationale for its choice.
To assess performance, we compute accuracy (\oursfour) or $F_1$(\oursthree). To this, we define Precision and Recall as in equation~\ref{equ:pre} and equation~\ref{equ:recall}.
\begin{equation}
\label{equ:pre}
\text{Precision} := \frac{n_o}{n_o + n_x}
\end{equation}
\begin{equation}
\label{equ:recall}
\text{Recall}  := \frac{n_o + n_x}{n_o + n_x + n_u}
\end{equation}
Where \( n_o \) denotes the number of correct answers, \( n_x \) the number of incorrect answers, 
and \( n_u \) the number of cases in which the model refused to answer (i.e., selected the “don’t know” option). 
Precision measures the accuracy among the answered questions, whereas Recall measures 
the proportion of answered questions relative to the total.

\section{Result and Analysis}

\subsection{Multiple-Choice Taxation Questions}
\subsubsection{Performance of LLMs on \oursthree}
In \oursthree, a model needs to decide whether the imposition of additional penalty tax is legitimate, based on provided facts and claims from both the plaintiff (taxpayer) and the defendant (tax authority) 
(Table \ref{tbl_data_example_eng} in Appendix \ref{emc}) .
We evaluate ten LLMs and permit them to refuse to answer if they are not confident. (Table \ref{tab:base_model_f1_plat_3mc}).  The results show that except for Exaone3.5 and GPT-4.1, all models show comparable $F_1$ scores 0.70-0.79 (col 1), with the commercial reasoning model o3 and Gemini2.5-pro achieving the highest score 0.79 $F_1$.
Interestingly, non-reasoning models (col 3, rows 1-6) tend to exhibit lower recall compared to reasoning models, suggesting that they are more likely to refrain from making a decision.\footnote{Motivated by the difference in the  number of “don’t know” responses between reasoning and non-reasoning models, we conducted a qualitative analysis. The non-reasoning models often provide logically sound (and sometimes even correct) explanation process well but they eventually conclude with statements such as “However, since the exact determination cannot be made,” or “Yet, because the tax authority’s criteria are often stringent, I answer “don’t know”.”)} In contrast, they generally achieve higher precision, indicating that when they do respond, their answers are more often correct.
These findings align with recent works indicating that reasoning models show similar or even lower performance on complex tasks \citep{shojaee2025illusionthinkingunderstandingstrengths}, and tend to exhibit overconfidence \citep{mei2025reasoning_trace_overconfident}.

\subsubsection{Cases LLMs Cannot Effectively Handle}
 To gain insight into what aspects LLMs are (not) capable of, 
 we manually analyzed cases where either the majority of LLMs answered correctly or incorrectly.\footnote{We conducted exploratory analysis with GPT-4o, GPT-o1-mini, Claude-3.5-sonnet, o1, EXAONE-3.0-7.8B-Instruct, and Qwen2.5-7B-Instruct}
LLMs were able to recognize the following principles:
 \begin{itemize}
    \item Ignorance or misunderstanding of tax laws by a taxpayer does not constitute a justifiable reason.\footnote{Daegu District Court 2015Guhap877}
    \item Mistakes or misunderstandings by tax accountants do not exempt taxpayers from responsibility; the final responsibility always lies with the taxpayer (thus, it is not a justifiable reason).\footnote{Seoul Administrative Court 2016Guhap56936}
 \end{itemize}
 On the other hand, LLMs show the following failure patterns. (Examples of failure patterns are in Appendix \ref{errormode})
 \begin{itemize}
    \item When a taxpayer is misled due to the tax authorities’ opinion, LLMs were unable to make a clear decision due to a conflict with the principle of legitimate expectation.\footnote{Busan High Court 2016Nu11, Seoul High Court 2020Nu43946}
     \item When judges considered various taxpayer-specific circumstances, including the feasibility of fulfilling obligations, LLMs tended to strictly adhere to principles and rules.\footnote{Daegu District Court 2018Guhap20506}
 \end{itemize}

Based on these observations, we categorized them into two groups--\texttt{Easy} and \texttt{Hard}--according to observed reasoning difficulty.

\subsubsection{Case Categorization}
The \texttt{Easy} group consists of 36 cases where the issue can be clearly spotted and leads to a single normative conclusion based on existing legal rules or precedents as described below.
\begin{itemize}
\small
    \item Clerical errors or omissions that do not substantially affect the underlying tax amount are not subject to penalty taxes.
    \item When the tax authority issued an incorrect tax disposition that misled the taxpayer, a penalty on the delayed base tax is considered unlawful.
    \item Mere misunderstanding or ignorance of the law does not constitute a justifiable reason.
    \item Claiming ignorance of facts that the taxpayer could have reasonably known is not accepted as a justifiable reason.
    \item Even if a tax attorney, legal representative, or employee was involved in the filing process, the final legal responsibility lies with the taxpayer; thus, no justifiable reason is accepted.
\end{itemize}
The remaining 64 cases are classified as \texttt{Hard}.

The categorization reveals that LLMs generally perform well on \texttt{Easy} cases (Table \ref{tab:base_model_f1_plat_3mc}, col 4--6) where rigid application of rules is sufficient. 
However, they struggle with \texttt{Hard} cases that require flexible legal reasoning, case-specific consideration, or weighing of competing principles depending heavily on the specific factual context as shown below (col 7-9) as described in detail below.

\begin{itemize}
\small
    \item Cases where the taxpayer faced unavoidable circumstances that hindered payment -- these often depend on the judge’s perspective and discretion regarding the taxpayer's circumstance.
    \item Cases requiring proper assessment of whether differences among tax authorities indicate genuine divergence in interpretation and whether the tax law itself was ambiguous.
    \item Cases requiring assessment of whether the taxpayer, despite delayed payment, promptly fulfilled their obligations upon becoming aware and was otherwise compliant -- or, conversely, whether they neglected their duties and failed to exercise due care in tax compliance.
    \item Cases dealing with whether an official interpretation (e.g., from a tax officer or written inquiry response) qualifies as a public opinion.
\end{itemize}

This analysis suggests that all LLMs struggle with cases that lack clear reasoning patterns and require a more comprehensive evaluation of all relevant circumstances (not only of the statutes but also of the taxpayer’s circumstances) to reach a decision.

\subsubsection{Causes of Low Recall}
Non-reasoning models that do not explicitly employ inference-time scaling generally exhibit lower recall compared to reasoning models (Table \ref{tab:base_model_f1_plat_3mc} row 1-6 vs row 7-12). 
This results in a higher absolute number of “Don't know” labels overall.
To further investigate this behavior, we removed the “Don't know” option from \oursthree\ creating \ourstwo\ and measured the accuracy instead of $F_1$.

Notably, when “Don't know” options were removed and non-reasoning models were forced to choose between two candidates, the accuracy was lower than their original precision (Table \ref{tab:base_model_f1_plat_3mc} col 2, row 1-6 vs Table \ref{tab:base_model_accuracy_plat_2mc_4mc} col 1, row 1-6) except for Claude3.7-sonnet and Gemini2.5-flash. This suggests that many of the previously abstained (“Don't know”) cases were not simply uncertain but would likely have been incorrectly answered.


\begin{table}[ht]
  \caption{Accuracy comparison of vanilla LLMs on \ourstwo\ (2 options w/o reasons) and \oursfour\ (4 options w/ reasons). Their difference ($\Delta$) is shown at final column.}
  \label{tab:base_model_accuracy_plat_2mc_4mc}
  \centering
  \begin{threeparttable}
  \setlength{\tabcolsep}{5pt}
  \renewcommand{\arraystretch}{1.2}
  \small
    \begin{tabular}{l|c|c|c}
        \toprule 
        \textbf{Model} & \textbf{\ourstwo} & \textbf{\oursfour} & $\Delta$ \\
        \toprule 
        Exaone3.5-32B & 0.60 & 0.79 & 0.19 \\
        Qwen3-32B & 0.60 & 0.73 & 0.13 \\
        \midrule
        Claude3.7-sonnet & 0.67 & 0.84 & 0.17\\
        Gemini2.5-flash & 0.68 & 0.79 & 0.11 \\
        GPT-4o & 0.57 & 0.78 & 0.21 \\
        GPT-4.1 & 0.55 & 0.83 & 0.28 \\
        \midrule
        Qwen3-32B$^{\dagger}$ & $0.60_{\stddev{\pm0.05}}$ & $0.73_{\stddev{\pm0.02}}$ & 0.13 \\
        o3-mini & $0.53_{\stddev{\pm0.02}}$ & $0.66_{\stddev{\pm0.02}}$ & 0.13 \\
        o1 & $0.55_{\stddev{\pm0.01}}$ & $0.69_{\stddev{\pm0.02}}$ & 0.14 \\
        o3 & $0.62_{\stddev{\pm0.01}}$ & $0.77_{\stddev{\pm0.04}}$ & 0.15\\
        Gemini2.5-flash$^{\dagger\text{i}}$ & $0.61_{\stddev{\pm0.01}}$ & $0.72_{\stddev{\pm0.01}}$ & 0.11\\
        Gemini2.5-pro$^{\dagger\text{i}}$ & $0.66_{\stddev{\pm0.01}}$ & $0.80_{\stddev{\pm0.00}}$ & 0.14\\
        \bottomrule
    \end{tabular}
    \begin{tablenotes}
    \item $^\dagger$Reasoning mode     
    \item $^{\text{i}}$thinking-budget: 4096
    \end{tablenotes}

  \end{threeparttable}
\end{table}

Interestingly, while reasoning models are more likely to respond under uncertainty, their decisions are not always reliable when forced to choose, as reflected in their accuracy scores (Table \ref{tab:base_model_f1_plat_3mc} col 2, row 7-12 vs Table~\ref{tab:base_model_accuracy_plat_2mc_4mc}, col 1, row 7-12).

\subsubsection{Analysis under the IRAC Framework}
\label{rproblem}
To further investigate the low performance of LLMs on our task, we examined non-reasoning and reasoning models through the lens of the IRAC framework.
\begin{itemize}
    \item \textbf{I (Issue)}: Both models are generally able to identify the legal issue accurately. In many cases, they correctly articulated the core dispute and built their reasoning on it.
    
    \item \textbf{R (Rule)}: Upon examining the legal sources cited by the models, we found that many were either outdated (e.g., superseded by newer statutes) or unverifiable.
\end{itemize}

Motivated by these findings, we conducted additional experiments with RAG\footnote{Retrieval-Augmented Generation}.

Interestingly, LLMs show at most a small improvement, or even degradation when using RAG (Table \ref{tab:rag_model_accuracy_plat_3mc}).
There may be two potential explanations: (1) 
even when provided with the appropriate legal rules, LLMs may still struggle with the “Application” and “Conclusion” stages;
 (2) retrieving truly relevant legal documents remains challenging, as highlighted in recent studies \citep{Zheng2025legal_rag,hou2025naacl_clerc,park2025lrage}. 
 
 Given that our retrieval pool is relatively small (consisting of 100 precedents and 4,042 statutory articles), we focus first on hypothesis (1). To this end, we construct a new set of multiple-choice questions, where each option includes both a proposed answer (“Conclusion”) and its corresponding rationale (“Application”).

\begingroup
\setlength{\tabcolsep}{6pt}
\renewcommand{\arraystretch}{0.9}
\begin{table}[ht]
  \caption{RAG scores on \oursthree}
  \label{tab:rag_model_accuracy_plat_3mc}
  \centering
  \resizebox{\linewidth}{!}{
  \begin{threeparttable}
  \setlength{\tabcolsep}{4pt}
  \renewcommand{\arraystretch}{1.2}
  \small
    \begin{tabular}{l|c|c|c}
        \toprule 
        \textbf{Model} & \textbf{F1} & \textbf{P} & \textbf{R} \\
        \toprule 
        Qwen3-32B & 0.75 & 0.60 & 0.98 \\
        Qwen3-32B (RAG) & 0.77 & 0.62 & 1.00 \\
        \midrule
        Claude3.7-sonnet & 0.74 & 0.63  & 0.91 \\
        Claude3.7-sonnet (RAG) & 0.74 & 0.60  & 0.96 \\
        \midrule
        GPT-4.1 & 0.65 & 0.66 & 0.64 \\
        GPT-4.1 (RAG) & 0.64 & 0.57 & 0.76 \\
        \midrule
        \midrule
        Qwen3-32B$^\dagger$ & $0.72_{\stddev{\pm0.05}}$ & $0.57_{\stddev{\pm0.04}}$ & $0.96_{\stddev{\pm0.06}}$ \\
        Qwen3-32B$^\dagger$ (RAG) & $0.73_{\stddev{\pm0.02}}$ & $0.61_{\stddev{\pm0.05}}$ & $0.91_{\stddev{\pm0.14}}$\\
        \midrule
        o3$^\dagger$ & $0.79_{\stddev{\pm0.03}}$ & $0.65_{\stddev{\pm0.04}}$ & $1.00_{\stddev{\pm0.00}}$ \\
        o3$^\dagger$ (RAG) & $0.75_{\stddev{\pm0.01}}$ & $0.64_{\stddev{\pm0.01}}$ & $0.92_{\stddev{\pm0.04}}$\\
        \bottomrule
    \end{tabular}
    \begin{tablenotes}
    \item Column names without bracket are taken from Table~\ref{tab:base_model_f1_plat_3mc} for comparison with the results after adding RAG.
    \item $^\dagger$Reasoning mode. $^{\text{i}}$ thinking-budget: 4096
    \end{tablenotes}
  \end{threeparttable}
    }
\end{table}
\endgroup

\subsection{Multiple-Choice Questions with Answer Rationales}
We extend the binary “lawful” and “unlawful” options from \ourstwo\ to a set of four answer choices--two labeled as “lawful” and two as “unlawful”--each accompanied by annotated rationales. These rationales are plausible but not necessarily correct (Table \ref{tbl_4_choices} in Appendix \ref{emc4}).

In the resulting benchmark, \oursfour, LLMs achieve higher accuracy scores (+ 0.13--0.28, Table \ref{tab:base_model_accuracy_plat_2mc_4mc}, col 2) despite the increased difficulty of selecting from four options (compared to the expected baseline accuracy of 0.50 for \ourstwo\space and 0.25 for \oursfour\space under random guessing).

This result suggests that LLMs struggle in the absence of guidance for the “Application” and “Conclusion” stages. It also implies that reasoning from the conclusion--i.e., backward chaining--may be beneficial in legal domains, though further investigation is needed. \footnote{We verified that this interpretation is not confounded by superficial shortcuts (Details are in \ref{superficial_clue_exp}).}
\cite{Poole_Mackworth_2023,zhou2023leasttomost,kazemi2023acl_lambada,lee2025naccl_symba}.

Notably, non-reasoning models (rows 1–6) show a larger improvement in accuracy compared to reasoning models (rows 6–9), suggesting that hints embedded in plausible rationales provide greater leverage for less capable models. To further analysis, we develop an essay-type benchmark.

\begingroup
\setlength{\tabcolsep}{6pt}
\renewcommand{\arraystretch}{1}
\begin{table*}[!htbp]
  \caption{Agent scores on \oursthree~(See Appendix \ref{platex} for the prompt).}
\label{tab:rag_model_accuracy_plat_3mc_agentic}
  \centering
    \small 

  \begin{threeparttable}
    \setlength{\tabcolsep}{3pt}%
    \renewcommand{\arraystretch}{0.95}%

    \begin{adjustbox}{max width=\textwidth}
    \begin{tabular}{@{}p{2.3cm}|c|c|c|c|c|c|c|c|c@{}}
      \toprule
      \textbf{Model} & \textbf{F1} & \textbf{P} & \textbf{R} & \textbf{F1-easy} & \textbf{P-easy} & \textbf{R-easy} & \textbf{F1-hard} & \textbf{P-hard} & \textbf{R-hard}\\
      \toprule
        GPT-4.1& 0.65 & 0.66 & 0.64
        &0.75 &0.76 &0.74 &0.49 & 0.42 &0.64 \\
        o3 & $0.79_{\stddev{\pm0.03}}$ & $0.65_{\stddev{\pm0.04}}$ & $1.00_{\stddev{\pm0.00}}$ & $0.83_{\stddev{\pm0.02}}$ & $0.71_{\stddev{\pm0.03}}$ & $1_{\stddev{\pm0.00}}$
            &$0.77_{\stddev{\pm0.05}}$&$0.62_{\stddev{\pm0.08}}$& $1_{\stddev{\pm0.00}}$ \\
        \midrule
        GPT-4.1 (RAG)& 0.64 & 0.57 & 0.76 & 0.62 & 0.56 & 0.74 & 0.66 & 0.57 & 0.77\\
        o3 (RAG) & $0.75_{\stddev{\pm0.01}}$ & $0.64_{\stddev{\pm0.01}}$ & $0.92_{\stddev{\pm0.04}}$ & $0.85_{\stddev{\pm0.03}}$ & $0.80_{\stddev{\pm0.05}}$ & $0.91_{\stddev{\pm0.03}}$ & $0.69_{\stddev{\pm0.02}}$ & $0.55_{\stddev{\pm0.01}}$ & $0.93_{\stddev{\pm0.05}}$\\
        \midrule
        GPT-4.1 (ReAct) & 0.76 & 0.63 & 0.97 & 0.74 & 0.60 & 0.97 & 0.77 & 0.64 & 0.96 \\
        o3 (ReAct) & $0.78_{\stddev{\pm0.02}}$ & $0.64_{\stddev{\pm0.02}}$ & $0.99_{\stddev{\pm0.01}}$ & $0.76_{\stddev{\pm0.08}}$ & $0.62_{\stddev{\pm0.10}}$ & $0.97_{\stddev{\pm0.02}}$ & $0.77_{\stddev{\pm0.04}}$ & $0.63_{\stddev{\pm0.05}}$ & $0.99_{\stddev{\pm0.01}}$ \\
        \bottomrule
    \end{tabular}
    \end{adjustbox}
  \end{threeparttable}
\end{table*}
\endgroup
\begin{table}[!htbp]
  \caption{Essay Score comparison across LLMs}
  \label{tab:baselines_acc_essay}
  \centering
  \begin{threeparttable}
  \setlength{\tabcolsep}{5pt}
  \renewcommand{\arraystretch}{1.2}
  \scriptsize
  \begin{tabular}{l|c|l|c}
    \toprule 
    \textbf{Model} & \textbf{Rubric Score} & \textbf{Model} & \textbf{Rubric Score}\\
    \midrule 
    Exaone-3.5-32B & 0.43 & Qwen3-32B$^\dagger$ & $0.45_{\stddev{\pm0.00}}$\\
    Qwen3-32B  & 0.48 & o3-mini  & $0.48_{\stddev{\pm0.01}}$ \\  
    Claude3.7-sonnet  & 0.62 & o1  & $0.59_{\stddev{\pm0.01}}$ \\  
    Gemini2.5-flash & 0.68 & o3  & $0.66_{\stddev{\pm0.01}}$ \\ 
    GPT-4o & 0.61 & Gemini2.5-flash$^{\dagger, \text{i}}$ & $0.70_{\stddev{\pm0.01}}$ \\ 
    GPT-4.1 & 0.65 & \textbf{Gemini2.5-pro}$^{\dagger, \text{i}}$ & $\textbf{0.71}_{\stddev{\pm0.01}}$ \\ 
    \bottomrule
    \end{tabular}
    \begin{tablenotes}
    \item $^\dagger$Reasoning mode
    \item $^{\text{i}}$ thinking-budget: 4096
    \end{tablenotes}

  \end{threeparttable}
\end{table}

\subsection{Essay-Type Questions}
For a comprehensive analysis, we construct \oursessay, which consists of 100 questions. For automatic evaluation using LLMs, we built ten task-specific rubrics made by tax professionals. Each rubric is aligned with the elements of IRAC (Table \ref{tbl_data_essay_IRAC_eng} in Appendix \ref{ee}). We assign 1 point for each rubric.

\subsubsection{Building a judge LLM}
 In essay-type examinations, scoring results can vary across graders even with the presence of rubrics~\citep{smolentzov2013automated, ART002023486, doi-etal-2024-automated}. Thus, it is standard to employ multiple graders in major essay-based examinations, including legal qualification exams. 
Following this practice, we conducted a human evaluation by asking three tax experts\footnote{Authors of this paper. One PhD student and two professors in Taxation Science.} to grade a random sample of 30 responses generated by \texttt{o3}. 

Next, to build a LLM judge, we benchmarked multiple LLMs by measuring their correlation with these expert averages. Among them, \texttt{o3} achieved the highest correlation and was therefore selected as the judge model (Pearson $r$ = 0.67, Spearman $\rho$ = 0.70, Figure~\ref{fig:avg_corr} in Appendix). The inter-human Pearson correlation scores are $r=0.79$, $r=0.53$, and $r=0.58$ respectively.

Using o3 as the judge, we conducted further analysis. We found that (1) o3 tends to assign more generous scores compared to human graders (although there is a rank-wise correlation between them), and (2) o3 cannot judge “Rule”-related rubrics where it needs to verify the cited legal documents.\footnote{Consistently, the responses from o3 achieves zero-score 91\% times on “Rule”-related rubrics when judged by human experts. This is in line with recent study from \citet{dahl2024largelegalfiction}}
Based on this observation, the four “Rule”-related rubrics were excluded and the maximum score per question was set to 6. (See Appendix \ref{llmasajudge} for more details.)

\subsubsection{Result}
Gemini2.5-pro achieves the top score (0.71) indicating that its generated answers, on average, satisfy approximately 4.25 out of 6 rubrics (Table \ref{tab:baselines_acc_essay}, bottom). Notably, reasoning models do not show clearly superior performance raising questions about the credibility of reasoning process on complex real-world tasks  \cite{shojaee2025illusionthinkingunderstandingstrengths,mei2025reasoning_trace_overconfident}.

\subsection{Agent-Based Approach}
\subsubsection{Motivation}
The experiments with \oursthree\space and \oursfour\space reveal that LLMs struggle with the “Application” and “Conclusion” stages. 
Another experiment with \oursessay, together with the qualitative analyses of the \oursthree\space\footnote{Section \ref{rproblem}}, reveals vanilla LLMs cannot handle “Rule” stage often citing outdated or unverifiable statutes and precedent. Based on these observation, we introduce agentic retrieval.

\subsubsection{Design}
We use the ReAct prompt \citep{yao2023react} that consists of three main steps: thought, action, and observation. We hypothesize that this  setup will benefit the “Rule” stage by iteratively refining retrieved legal documents. In this regard, we enforce a minimum of three retrieval steps.

For “action” steps, we introduce three tools: a BM25 retriever, a legal-document comparison tool, and a legal-analyzer tool. The legal document-relevance-analyzer tool classifies and summarizes relevant legal documents while the legal-analyzer tool to enables LLMs to reason following IRAC. Details for agentic retrieval are in Appendix \ref{additionaltools}.



\subsubsection{Result \& Analysis}
With the agentic-based approach, GPT-4.1 achieves +0.11 F\textsubscript{1} (row 1 vs 5, col 1) while o3 shows -0.01 F\textsubscript{1}. In both cases, the agentic retrieval demonstrates superior performance compared to RAG (+0.12 F$_1$ for GPT-4.1 and +0.03 F$_1$ for o3).


Further analysis reveals, under ReAct, the models initially retrieved only precedents, but gradually  began incorporating statutes into the retrieval process; after the 3rd retrieval, GPT-4.1 retrieves documents comprising 56\% precedents and 44\%  statutes while o3 retrieves 66\%  precedents and 34\% statutes.

This suggests that under single-pass RAG with a BM25 retriever, LLMs fail to access the relevant statutes needed for legal verification, resulting in low performance. 
However, with ReAct, the agent iteratively retrieves relevant statutes and incorporates the retrieved knowledge into the reasoning process, leading to higher recall (col 3, row 5--6). 
Precision remains similar to that of vanilla LLMs, indicating that the agentic approach also cannot significantly enhance the “Application” and “Conclusion” steps.

Interestingly, o3 does not exhibit significant improvements. This may be because o3 already incorporates inference-time reasoning strategies, reducing the marginal benefit of additional prompting or agentic structuring.



\section{Conclusion}
We introduce \ours, a benchmark designed to evaluate LLMs' capability in taxation. Compared to previous studies, our dataset includes cases where answers cannot be determined solely by referencing statutes, requiring a deeper understanding of legal and contextual factors of individual legal issues.
Our experiments reveal that while LLMs demonstrate some capability, vanilla models struggle to comprehensively understand taxation issues, particularly at the higher reasoning stages (Application/Conclusion of IRAC).
We also show that by gradually integrating retrieval and self-reasoning, these limitations can be partially mitigated, although reaching a correct conclusion remains challenging.

\section{Limitation}
Tax accountants require a broad range of knowledge and advanced reasoning skills. For instance, the Korean Certified Tax Accountant (CTA) exam, a professional qualification for tax practitioners, covers multiple subjects: multiple-choice exams in Public Finance, Introduction to Tax Law, and Introduction to Accounting; written exams in Tax Law I (covering Corporate Tax Law, Income Tax Law, etc.) and Tax Law II (covering Value-Added Tax Law, Inheritance and Gift Tax Law, etc.).
On the other hand, our study, however, focuses specifically on evaluating the justifiability of exemption from additional tax penalties, serving as a case study in which LLMs must demonstrate a comprehensive understanding of complex situations rather than simply referencing tax law statutes. Since our study is grounded in Korean tax law, care should be taken when extending these findings across different languages, jurisdictions, or legal domains.

A more holistic evaluation of LLMs in the tax domain remains as a future work.

\section*{Acknowledgments}
Eunkyung Choi, Hongseok Oh, Juheon Kang, and Wonseok Hwang were supported by the National Research Foundation of Korea (NRF) grant funded by the Korea government(MSIT) (RS-2025-23524855). Eunkyung Choi and Juheon Kang were also partially supported by the National Research Foundation of Korea(NRF) grant funded by the korea government(MSIT) (NO. RS-2022-NR068754).

\bibliography{legal_ai}
\newpage
\twocolumn
\appendix

\section{Detailed statistics about \ours}
\label{detailedstatistics}
Detailed statistics about \ours\space are presented in Table \ref{tab:data_stat}. \ours\space consists of 36 easy cases and 64 hard cases.

\begingroup
\section{Example of \ours}
\label{platex}
\subsection{\oursthree}
\label{emc}
Examples from \oursthree\space are presented in Table \ref{tbl_data_example_eng}. 
\subsection{\oursfour}
\label{emc4}
Examples from \oursfour\space are presented in Table \ref{tbl_4_choices}. 
\subsection{\oursessay}
\label{ee}
Essay questions are based on \oursthree\space with prefix that follows the format of the second-round essay-style exam for the Korean Certified Tax Accountant (CTA): “Read the following case and answer the question. Explain whether the imposition of additional tax (penalty tax) in the above case is lawful.
(Note: Only the lawfulness of the imposition of additional tax is to be considered; no other issues are to be taken into account.)\\
<Case>”\\
Examples from \oursessay\space rubrics are presented in Table \ref{tbl_data_essay_IRAC_eng}. 
\endgroup
\begingroup
\section{Dataset Annotation Procedure}
\label{datasetannotaion}
For \oursthree, one tax expert annotated 100 precedents (hours). Other tax expert subsequently reviewed all annotated cases to ensure consistency and accuracy.

For \oursfour, one tax expert annotated 100 precedents (22 hours) and another tax expert reviewed all the annotated precedents along with the associated multiple-choice options, revising 8 of them. This collaborative review process enabled the construction of a high-quality dataset.

For \oursessay, three tax experts curated IRAC-grounded scoring rubrics by consensus.

All datasets were annotated in accordance with the Court’s legal reasoning and the Judge’s final decisions. The annotation GUI are shown in Figure~\ref{fig:labelstudio} and Figure~\ref{fig:labelstudio2}.

\section{Prompt to Answer \ours}
\subsection{Prompt to Answer \oursthree\space and \oursthree(w RAG)}
Prompts to Answer \oursthree\space and \oursthree(w RAG) are presented in Figure \ref{promptmc}. 
Korean is translated to English using GPT-4o.

\subsection{Prompt to Answer \oursfour\space}
Prompt to Answer \oursfour\space is shown in Figure \ref{promptplatmc4}. 
\subsection{Prompt to Answer \oursessay\space}
Prompt to Answer \oursessay\space is shown in Figure \ref{promptessay} and prompt for LLM-as-a-Judge scoring is in Figure \ref{llm_judge_prompt}.  These prompts were designed based on the answer and grade prompts introduced by \citet{fan2025lexambenchmarkinglegalreasoning}.
\endgroup

\begingroup
\section{Examples of Models common failure patterns}
\label{errormode}
\begin{itemize}
    \item When a taxpayer is misled due to the tax authorities’ opinion, LLMs were unable to make a clear decision due to a conflict with the principle of legitimate expectation. (Worked examples are in Figure \ref{workedexfailurepatterns-case1}.)
    \item When judges considered various taxpayer specific circumstances, including the feasibility of fulfilling obligations, LLMs tended to strictly adhere to principles and rules. (Worked examples are in Figure \ref{workedexfailurepatterns-case2}.)
\end{itemize}
\endgroup

\begingroup
\section{Analysis of Performance Gains in \oursfour: Examining Superficial Cues
}
\label{superficial_clue_exp}
We examined whether performance gains on \oursfour\ might be confounded by superficial shortcuts, focusing on lexical overlap and length-based cues.

\paragraph{Lexical Overlap Analysis}
We measured lexical overlap between each question and its answer options (both correct and incorrect) using ROUGE-1, ROUGE-2, and ROUGE-L F1 scores. The overlap scores are consistently low (0.08–0.12), indicating minimal n-gram sharing between the input and the options. ROUGE-1 F1 scores are identical (0.12) for both correct and incorrect options. Incorrect options display marginally higher overlap scores across ROUGE-2 F1 (0.09 vs. 0.08) and ROUGE-L F1 (0.10 vs. 0.09), suggesting that incorrect options may be slightly more attractive at the superficial lexical level.

This aligns with our design goal: when constructing distractors, we avoided trivial edits such as changing a number or swapping a single word (e.g., from ``financial income'' to ``earned income''). Instead, we attempted to craft attractive but incorrect options that require assessing whether the underlying circumstance supports the legal conclusion.

\paragraph{Length-Based Cue Analysis}
We also examined length-based cues. Correct options average 87 ($\pm$25.68) tokens, while incorrect options average 78 ($\pm$21.25) tokens. The Pearson correlation between correct option length and mean distractor length is $r = 0.1287$ ($p = 0.202$).

These observations do not provide evidence that superficial shortcuts drive performance, and are at least consistent with the possibility that backward-chaining--style reasoning may play a role. However, further investigation would be needed to confirm this interpretation.
\endgroup

\begingroup
\section{Details in Agentic Retrieval}
\paragraph{Legal Think Tools}
\label{additionaltools}
To further explore the legal agent methodology, we built additional tools that are employed during the agent experiments. They are commonly used in Agent-Based Approach setting.
\begin{itemize}
    \item \textbf{Bridge between “Rule” and “Application”}\\
    \textit{Document-Relevance-Analyzer Tool}: summarizes statutes and precedents, classifies relevant vs.\ irrelevant parts, and compares them with the issue (Figure \ref{promptmadc}).
    \item \textbf{Precursor to IRAC structured reasoning}\\
    \textit{Legal-Analyzer Tool}: implements a structured IRAC analysis; identifies the key issue, states the governing rules, applies them to the facts, and concludes with the final legal determination (Figure \ref{promptmala}).
\end{itemize}
\paragraph{Worked example in Agentic Retrieval}
An worked example is in Figure \ref{workedexampleinagenticretrieval}.
\endgroup

\begingroup
\section{Additional Agent experiment}

\setlength{\tabcolsep}{6pt}
\renewcommand{\arraystretch}{1}
\begin{table*}[t]
  \caption{Agent scores on \oursthree\space (M stands for the experiment with multiple agents with different roles.)}
\label{tab:rag_model_accuracy_plat_3mc_multi_agentic}
  \centering
    \small 

  \begin{threeparttable}
    \setlength{\tabcolsep}{3pt}%
    \renewcommand{\arraystretch}{0.95}%

    \begin{adjustbox}{max width=\textwidth}
    \begin{tabular}{@{}p{2.6cm}|c|c|c|c|c|c|c|c|c@{}}
      \toprule
      \textbf{Model} & \textbf{F1} & \textbf{P} & \textbf{R} & \textbf{F1-easy} & \textbf{P-easy} & \textbf{R-easy} & \textbf{F1-hard} & \textbf{P-hard} & \textbf{R-hard}\\
      \toprule
        GPT-4.1& 0.65 & 0.66 & 0.64
        &0.75 &0.76 &0.74 &0.49 & 0.42 &0.64 \\
        o3 & $0.79_{\stddev{\pm0.03}}$ & $0.65_{\stddev{\pm0.04}}$ & $1.00_{\stddev{\pm0.00}}$ & $0.83_{\stddev{\pm0.02}}$ & $0.71_{\stddev{\pm0.03}}$ & $1_{\stddev{\pm0.00}}$
            &$0.77_{\stddev{\pm0.05}}$&$0.62_{\stddev{\pm0.08}}$& $1_{\stddev{\pm0.00}}$ \\
        \midrule
        GPT-4.1 (RAG)& 0.64 & 0.57 & 0.76 & 0.62 & 0.56 & 0.74 & 0.66 & 0.57 & 0.77\\
        o3 (RAG) & $0.75_{\stddev{\pm0.01}}$ & $0.64_{\stddev{\pm0.01}}$ & $0.92_{\stddev{\pm0.04}}$ & $0.85_{\stddev{\pm0.03}}$ & $0.80_{\stddev{\pm0.05}}$ & $0.91_{\stddev{\pm0.03}}$ & $0.69_{\stddev{\pm0.02}}$ & $0.55_{\stddev{\pm0.01}}$ & $0.93_{\stddev{\pm0.05}}$\\
        \midrule
        GPT-4.1 (ReAct) & 0.76 & 0.63 & 0.97 & 0.74 & 0.60 & 0.97 & 0.77 & 0.64 & 0.96 \\
        o3 (ReAct) & $0.78_{\stddev{\pm0.02}}$ & $0.64_{\stddev{\pm0.02}}$ & $0.99_{\stddev{\pm0.01}}$ & $0.76_{\stddev{\pm0.08}}$ & $0.62_{\stddev{\pm0.10}}$ & $0.97_{\stddev{\pm0.02}}$ & $0.77_{\stddev{\pm0.04}}$ & $0.63_{\stddev{\pm0.05}}$ & $0.99_{\stddev{\pm0.01}}$ \\
        \midrule 
        GPT-4.1 (ReAct+M) & 0.75 & 0.61 & 0.99 & 0.81 & 0.70 & 0.97 & 0.71 & 0.56 & 0.99 \\
        o3 (ReAct+M) & $0.73_{\stddev{\pm0.01}}$ & $0.59_{\stddev{\pm0.00}}$ & $0.96_{\stddev{\pm0.03}}$ & $0.80(_{\stddev{\pm0.02}}$ & $0.67_{\stddev{\pm0.01}}$ & $1.00_{\stddev{\pm0.00}}$ & $0.71_{\stddev{\pm0.01}}$ & $0.56_{\stddev{\pm0.01}}$ & $0.98_{\stddev{\pm0.01}}$ \\
        \bottomrule
    \end{tabular}
    \end{adjustbox}
  \end{threeparttable}
\end{table*}

\subsection{Multi-agent collaboration}
To further investigate the legal agent methodology, we employ multi-agent collaboration for the LLM to respond to \ours.

\subsubsection{Experiments}
For the purpose of our experiments, we constructed the Virtual-Court Tool. Virtual-Court Tool is a role-playing tool where the plaintiff’s lawyer argues against the penalty tax, the defendant’s lawyer argues for it, and the judge neutrally compares both arguments to reach a lawful conclusion (Figure \ref{promptvc})

\subsubsection{Result}
In our multi-agent collaboration experiments, three LLMs are assigned specific roles: an attorney for the taxpayer, an attorney for the tax authority, and a judge. We hypothesize that this setup can benefit the ``Application'' stage by enforcing diverse legal perspectives, thereby bypassing the need to determine which viewpoint is lawful during inference. The result is shown at Table \ref{tab:rag_model_accuracy_plat_3mc_multi_agentic}.

Both models, GPT-4.1 and o3, show a slight performance drop when combined with ReAct prompting and multi-agent collaboration (bottom row). 
For easy cases, both GPT-4.1 and o3 achieve higher precision with ReAct + multi-agent than with ReAct alone (col 5, row 3 vs 4); recall remains unchanged for GPT-4.1 and increases for o3 (col 6, row 3 vs 4). For hard cases, in contrast, both models show notably lower precision with ReAct + multi-agent (col 8, row 3 vs 4); recall is higher for GPT-4.1 and remains similar for o3 (col 9, row 3 vs 4). Overall, lower precision in hard cases drives a decrease in F1.

\paragraph{Analysis}
We further analyzed how the number of correct, incorrect, and “don’t know” responses changed for individual categories (Table \ref{tab:data_stat}).

GPT-4.1 improved in most easy categories, especially in \texttt{easy-(C)} and \texttt{easy-(D)} (Table~\ref{reactMcomparison41}).
Similarly, o3 improved in most easy categories, however, unlike GPT-4.1, it degraded in \texttt{easy-(C)} (Table \ref{reactMcomparisono3}).

For hard cases, both GPT-4.1 and o3 improved in \texttt{hard-(I)}, whereas GPT-4.1 degraded in \texttt{hard-(F)} and \texttt{hard-(G)} (Table~\ref{reactMcomparison41}). Similarly, o3 also degraded in \texttt{hard-(F)} (Table~\ref{reactMcomparisono3}).

Although we conducted a qualitative analysis as above, we did not observe clear patterns. Building successful Multi-agent ReAct agents remains as a future study.

\begin{table}[!htbp]
\centering
\caption{Changes in model answers across categories (GPT-4.1 ReAct to ReAct+M).}
\resizebox{0.9\linewidth}{!}{
\begin{tabular}{lccc}
\hline
\textbf{Categories} & $\Delta$ \textbf{Correct} & $\Delta$ \textbf{Incorrect} & $\Delta$ “Don't know” \\
\hline
easy-(A) & $-1$ & $+1$ & $0$ \\
easy-(B) & $+1$ & $-1$ & $0$ \\
\textbf{easy-(C)} & \textbf{+2} & $-1$ & $-1$ \\
\textbf{easy-(D)} & \textbf{+2} & $-2$ & $0$ \\
easy-(E) & $+1$ & $-1$ & $0$ \\
\textbf{hard-(F)} & $-1$ & \textbf{+2} & $-1$ \\
\textbf{hard-(G)} & $-2$ & \textbf{+3} & $-1$ \\
\textbf{hard-(H)} & \textbf{+3} & $-2$ & $-1$ \\
hard-(I) & \textbf{+2} & $0$ & $-2$ \\
\hline
\end{tabular}
}
\label{reactMcomparison41}
\end{table}

\begin{table}[!htbp]
\centering
\caption{Changes in model answers across categories (o3 ReAct to ReAct+M).}
\resizebox{0.9\linewidth}{!}{
\begin{tabular}{lccc}
\hline
\textbf{Categories} & $\Delta$ \textbf{Correct} & $\Delta$ \textbf{Incorrect} & $\Delta$ \textbf{Don't know}\\
\hline
easy-(A) & $0$ & $0$ & $0$ \\
easy-(B) & $+1$ & $-1$ & $0$ \\
\textbf{easy-(C)} & $-2$ & \textbf{+2} & $0$ \\
easy-(D) & $+1$ & $-1$ & $0$ \\
easy-(E) & $+1$ & $-1$ & $0$ \\
\textbf{hard-(F)} & $-3$ & \textbf{+3} & $0$ \\
hard-(G) & $-1$ & $0$ & $1$ \\
hard-(H) & $+1$ & $0$ & $-1$ \\
\textbf{hard-(I)} & \textbf{+2} & $-2$ & $0$ \\
\hline
\end{tabular}
}
\label{reactMcomparisono3}
\end{table}


\section{LLM-as-a-judge for essay grader}
\label{llmasajudge}
To refine LLM grader, we conduct rubric-level analyses comparing human graders’ scores with \texttt{o3}’s scores. 
Our observations are as follows:
\begin{itemize}
\item Compared to human graders, the LLM-based grader consistently assigns more generous scores across rubrics.
\item Writer LLMs, including the state-of-the-art reasoning model \texttt{o3}, frequently hallucinate in the Rule (R) component of IRAC.
\end{itemize}
\subsection{Generosity Bias of LLM Graders}
Figure \ref{red_scatter3} presents scatter plots comparing the LLM-based grader (o3) with each of the three human graders. Across all graders, the LLM’s scores exhibit a positive correlation with human judgments, although the strength of correlation varies substantially: Pearson’s r ranges from 0.44 with human1 to 0.76 with human3. The fitted regression lines show that o3 generally gives higher scores than the human graders, as reflected in slopes less than one. This suggests that, while the LLM captures grading trends consistent with human judgment, it consistently produces more generous assessments.
\subsection{Hallucination in the Rule (R) Component}
Human graders assigned almost zero points across all R-based rubrics for the \texttt{o3} responses. A qualitative analysis revealed that these responses exhibited severe hallucinations, consistent with the issues discussed in Sec.~\ref{rproblem}.\\ 
\citet{dahl2024largelegalfiction} shows that while more recent large language models demonstrate reduced hallucination rates when evaluated on legal texts and case law, the problem remains significant. Building on this insight, our study pursued more precise scoring by excluding the four rubrics classified under R. (No 2, 4, 6, 8 rubrics. You can check them in Table \ref{tbl_data_essay_IRAC_eng})
\endgroup

\section{Resources and Licensing}
Datasets and models used in this work, with licenses are presented in Table \ref{license}. Additionally, \ours\ was preprocessed to anonymize all personally identifiable information (PII), despite the dataset’s origin in court precedents.

\begingroup
\begin{figure*}[t]
    \centering
    \begin{subfigure}{0.45\linewidth}
        \centering
        \includegraphics[width=\linewidth]{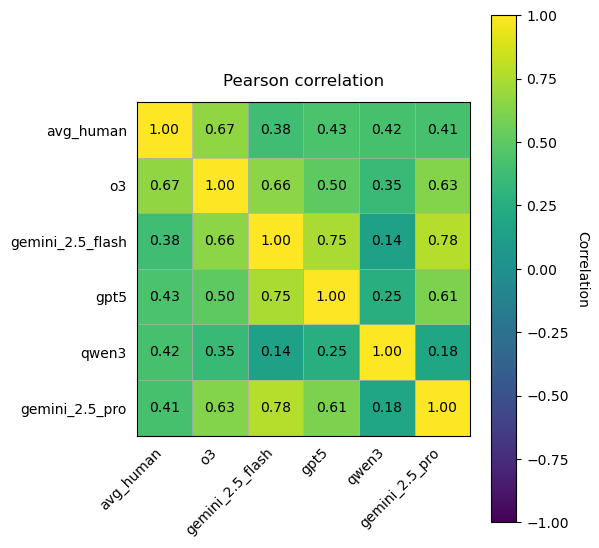}
        \caption{Pearson correlation}
    \end{subfigure}
    \hfill
    \begin{subfigure}{0.45\linewidth}
        \centering
        \includegraphics[width=\linewidth]{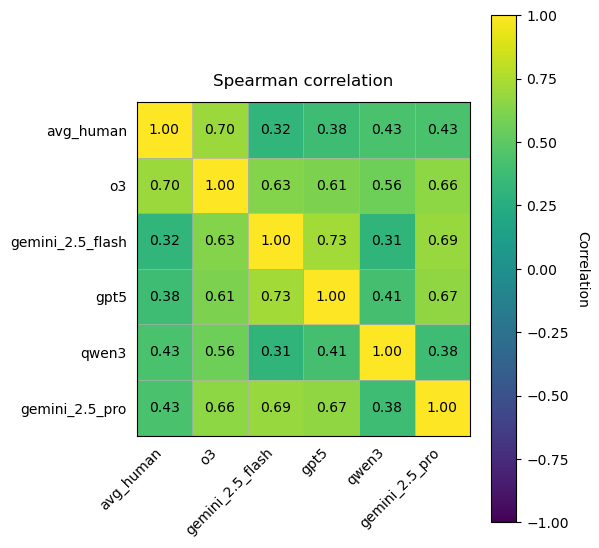}
        \caption{Spearman correlation}
    \end{subfigure}
    
    \caption{Pearson and Spearman correlations between candidate LLMs and expert average scores.}
    \label{fig:avg_corr}
\end{figure*}

\begingroup
\begin{table*}[!htbp]
\centering
\small
\setlength{\tabcolsep}{6pt}
\renewcommand{\arraystretch}{1.3}

\caption{Data Distribution by Year (2007--2023) and Difficulty-Based Topic Categorization for \ours}
\label{tab:data_stat}

\vspace{0.5em}
\text{(a) Data Distribution by Year (2007--2023)}
\vspace{0.3em}
\begin{tabular}{c ccccccccccccccccc}
\toprule
\textbf{Year} & 07 & 08 & 09 & 10 & 11 & 12 & 13 & 14 & 15 & 16 & 17 & 18 & 19 & 20 & 21 & 22 & 23 \\
\midrule
\textbf{Count} & 1 & 1 & 0 & 5 & 11 & 5 & 8 & 7 & 13 & 10 & 7 & 11 & 5 & 11 & 2 & 1 & 2 \\
\bottomrule
\end{tabular}

\vspace{1.2em}

\text{(b) Penalty Tax Justifiability Topics by Difficulty}

\vspace{0.3em}
\begin{tabularx}{\textwidth}{l X S}
\toprule
\textbf{Difficulty} & \textbf{Topic} & \textbf{Count} \\
\midrule
easy & (A) Clerical errors or omissions that do not substantially affect the underlying tax amount are not subject to penalty taxes. & 1 \\
easy & (B) When the tax authority issued an incorrect tax disposition that misled the taxpayer, a penalty on the delayed base tax is considered unlawful. & 3 \\
easy & (C) Mere misunderstanding or ignorance of the law does not constitute a justifiable reason. & 17 \\
easy & (D) Claiming ignorance of facts that the taxpayer could have reasonably known is not accepted as a justifiable reason. & 11 \\
easy & (E) Even if a tax attorney, legal representative, or employee was involved in the filing process, the final legal responsibility lies with the taxpayer; thus, no justifiable reason is accepted. & 4 \\
\midrule
hard & (F) Cases where the taxpayer faced unavoidable circumstances that hindered payment. & 22 \\
hard & (G) Cases requiring proper assessment of whether differences among tax authorities indicate genuine divergence in interpretation or whether the tax law itself was ambiguous. & 18 \\
hard & (H) Cases requiring assessment of whether the taxpayer, despite delayed payment, promptly fulfilled obligations--or neglected duties and failed to exercise due care. & 12 \\
hard & (I) Cases dealing with whether an official interpretation (e.g., from a tax officer or written inquiry response) qualifies as a public opinion. & 12 \\
\bottomrule
\end{tabularx}
\end{table*}

\setlength{\tabcolsep}{1pt}
\renewcommand{\arraystretch}{0.85}
\begin{table*}[!htbp]
\scriptsize
  \caption{
    Examples from \oursthree.}
\vspace{1em}
\vspace{1em}
  \label{tbl_data_example_eng}
  \centering
  \begin{threeparttable}
  \begin{tabular}{
   >{\raggedright\arraybackslash}p{4.8cm}
   |>{\raggedright\arraybackslash}p{4.8cm}
   |>{\raggedright\arraybackslash}p{4.8cm}
   |>{\raggedright\arraybackslash}p{1cm}
   }
    \toprule
    \textbf{Facts} 
    & \textbf{Claim from Plaintiff (Taxpayer)}
    & \textbf{Claim from Defendant (Tax Authority)}
    & \textbf{Label}
    \\
    \toprule 
        1. The plaintiff is a company established for the purpose of shipbuilding and sales. 
        
        2. On March 25, 2009, the plaintiff applied to the head of the Jungbu Tax Office for an extension of the payment deadline for KRW 120 billion out of KRW 145,381,546,613 in corporate tax for the 2008 tax year, and it was approved. 
        
        3. The plaintiff paid the remaining corporate tax of KRW 253,815,466.13, for which no extension application was filed, on March 31, 2009, and paid the inhabitant tax on corporate tax to the defendant on April 30, 2009. 
        
        4. On June 25, 2009, the plaintiff applied for an additional extension of the corporate tax for which the payment deadline had been extended, and the payment deadline was approved until September 30, 2009. 
        
        5. The plaintiff paid KRW 6,313,838,780 in inhabitant tax on corporate tax for the extended corporate tax payment deadline to the defendant on October 30, 2009. 
        
        6. The defendant imposed an additional tax of KRW 1,609,397,490, claiming that the plaintiff had not separately applied for an extension of the inhabitant tax payment deadline, even though it had received an extension of the corporate tax payment deadline. 
        
        7. After paying the additional tax, the additional tax was revised to KRW 1,105,805,430 according to the reduction decision. 
        8. The plaintiff applied for an exemption from the additional tax on February 5, 2010, but the defendant rejected it.

    &
     Plaintiff's Arguments and Grounds 1.
     
     Illegality of Imposing Penalty Tax Due to Justifiable Cause 
        
        - The plaintiff received approval for an extension of the corporate tax payment deadline, and therefore mistakenly believed that the resident tax payment deadline was also extended accordingly. 
        
        - The defendant's staff member answered that the resident tax payment deadline would be extended, leading the plaintiff to believe this. - Therefore, the plaintiff had a justifiable reason for failing to pay the resident tax within the deadline. 
        
        - Relevant Laws: Framework Act on National Taxes Article 6, Local Tax Act Article 27-2 Paragraph 
        
    2. Penalty Tax Exemption is a Mandatory Act and Meets the Exemption Requirements 
    
        - The plaintiff was facing a significant crisis in its business, which constitutes a reason for penalty tax exemption. 
        
        - The defendant has an obligation to accept the exemption application. - Relevant Laws: Local Tax Act Article 27-2 Paragraph 2, Enforcement Decree Article 13-2, Article 11 Paragraph 1 Item 4 

    &
    
    Defendant's Arguments and Grounds 1.
    
    Illegality of Penalty Tax Exemption Application 
    
        - The application for penalty tax exemption must be made by the end of the statutory payment deadline. 
        
        - The plaintiff applied on February 5, 2010, past the statutory payment deadline, making it an illegal application. 
    
        - Relevant Laws: Local Tax Act and related Enforcement Decree 
    
    2. Non-Existence of Justifiable Cause 
    
        - The extension of the corporate tax payment deadline is irrelevant to the extension of the resident tax payment deadline. 
        
        - The plaintiff merely misunderstood the laws and administrative interpretations, and there was no response from the defendant's staff member. 
        
        - Therefore, the plaintiff has no justifiable reason. 
    
    3. Does Not Fall Under the Penalty Tax Exemption Requirements 
    
        - The plaintiff does not fall under the penalty tax exemption requirement of “when the business is in a significant crisis”. 
        
        - Therefore, the rejection of the exemption application is lawful. 
        
        - Relevant Laws: Local Tax Act Enforcement Decree Article 11 Paragraph 1 Item 4
    & Not legitimate. \\
    \toprule
    1. The plaintiff operated a charging station from 2011 and opened and reported a business account. 
    
    2. In 2013, the plaintiff's revenue exceeded 300 million won, making them obligated to use double-entry bookkeeping from January 1, 2014 (Article 160, Paragraph 3 of the former Income Tax Act; Article 208, Paragraph 5, Subparagraph 2 of the former Enforcement Decree of the Income Tax Act). 
    
    3. The plaintiff newly opened this charging station on April 1, 2014, and opened five deposit accounts (hereinafter referred to as 'the deposit accounts in this case') at NongHyup Bank, but did not report the opening of a business account to the competent tax office by June 30, 2015 (Article 160-5, Paragraph 3 of the former Income Tax Act). 
    
    4. In early May 2015, the plaintiff confirmed that 'Not applicable' was written in the 'Non-establishment of business account' item among the 'Penalty items' of the '2014 Comprehensive Income Tax Return Guidance' received from the defendant. 
    
    5. As a person subject to faithful reporting, the plaintiff reported the ending balance of some of the deposit accounts in this case when filing comprehensive income tax returns for 2014 and 2015.    
    &
    Plaintiff’s Arguments 1. 
    
    Existence of Justifiable Grounds: The plaintiff received a '2014 Tax Year Comprehensive Income Tax Notice' in May 2015, which stated 'Business Account Not Established' as 'Not Applicable.' 
    
    Therefore, the plaintiff believed that these deposit accounts had already been reported. 
    
    Thus, there are justifiable grounds for failing to fulfill the reporting obligation, and the plaintiff should be granted a reduction of penalties under Article 48 of the Framework Act on National Taxes. 
    
    2. Violation of the Principle of Taxation Based on Substance: The plaintiff actually used these deposit accounts as business accounts and fully listed the ending balance of the accounts in the faithful declaration confirmation form when filing comprehensive income tax. The defendant was able to understand the account details through this, but the defendant imposed penalties for the formal reason that it was not simply reported, which violates the principle of taxation based on substance in tax law. 
    &
    Defendant’s Arguments 1. Legality of Penalty Imposition: The defendant argues that the penalty imposition is lawful because the plaintiff, as a person obligated to use double-entry bookkeeping, did not fulfill the obligation to report business accounts under Article 160-5 of the former Income Tax Act.
    &
    Legitimate.
    \\
    \bottomrule
  \end{tabular}
  \end{threeparttable}
    \vspace{-5mm}
\end{table*}

\setlength{\tabcolsep}{4pt}
\renewcommand{\arraystretch}{1}
\begin{table*}[!htbp]
\scriptsize
  \caption{Example of \oursfour\space(The corresponding questions are given in Table \ref{tbl_data_example_eng}.)}
  \label{tbl_4_choices}
  \centering
  \vspace{-2mm}
  \begin{tabular}{c|p{11.5cm}}
   \toprule
    A & It is lawful to impose a penalty because an additional extension was requested for the extended deadline for corporation tax payment. \\
    \midrule
    B & Since approval was obtained from the Jungbu Regional Tax Office for an extension of the payment deadline for a portion of the corporate tax for the 2008 tax year, the imposition of penalties is lawful. \\
    \midrule
    C & Since the business is facing a significant crisis and falls under the grounds for exemption from additional tax, the imposition of additional tax is not lawful. \\
    \midrule
    D & It is not legitimate to impose a penalty tax because the company was established for the purpose of shipbuilding and sales. \\
    \midrule
    \midrule
    A & Opened a charging station and opened five deposit accounts (hereinafter 'the accounts in question') at Nonghyup Bank, but did not report the opening of business accounts to the competent tax office by June 30, 2015. Therefore, the imposition of the penalty tax is illegal. \\
    \midrule
    B & The imposition of a penalty tax is lawful in special circumstances where the taxpayer's unawareness of their obligation can be considered not unreasonable. \\
    \midrule
    C & The imposition of the additional tax is lawful. The 'Not Applicable' on the notice merely signifies that it is not subject to the additional tax for failure to open a business account, and does not imply that the account has already been reported. Therefore, the plaintiff's misunderstanding cannot constitute a valid reason. \\
    \midrule
    D & The plaintiff actually used each of the deposit accounts in question as a business account and, when filing the comprehensive income tax return, recorded all year-end balances of the accounts in the sincere filing confirmation form. Although the defendant could have ascertained the account details through this, imposing a penalty tax merely on the formal ground of non-reporting runs counter to the substantive taxation principle of tax law. Therefore, the imposition of the penalty tax is illegal. \\
    \bottomrule
  \end{tabular}
  \vspace{1em}
\end{table*}

\begin{table*}[!htbp]
\centering
\scriptsize
\caption{\oursessay~IRAC classification of PLAT-E rubrics}
\label{tbl_data_essay_IRAC_eng}
    \vspace{-2mm}
\begin{tabularx}{\textwidth}{l X S}
\toprule
\textbf{No} & \textbf{Rubric Items} & \textbf{IRAC element} \\
\midrule
1&Nature and Definition of Additional Tax (1 point):
Did the answer clearly describe the legal nature of additional tax (a sanction for violation of tax law obligations)?&I\\
2&Citation of Relevant Laws (1 point):
Did the answer accurately mention the relevant statutes such as the Framework Act on National Taxes, and correctly cite article numbers and provisions?&R\\
3&Requirements and Scope of Imposing Additional Tax (1 point):
Did the answer accurately explain the requirements for imposing additional tax and its scope of application in relation to the problem scenario?&I\\
4&Legal Basis for Exemption due to “Justifiable Reason” (1 point):
Did the answer correctly present the legal basis for exemption from additional tax (Article 48 of the Framework Act on National Taxes)?&R\\
5&Applicability of “Ignorance of Law,” etc. (1 point):
Did the answer clearly state that ignorance or misunderstanding of the law, or the lack of awareness by an expert, does not constitute a “justifiable reason”?&I\\
6&Accuracy of Relevant Case Law (1 point):
Was the cited case law an actual, existing case, and was its holding accurately identified?&R\\
7&Appropriateness of Case Holding Description (1 point):
Was the holding of the cited case clearly articulated and organically connected to the issue, contributing to the logical development of the argument?&A\\
8&Accuracy of Case Number and Source (1 point):
Was the case number (e.g., Supreme Court Decision 20XXDuXXXX) accurately indicated?&R\\
9&Identification of Core Issues and Introduction (1 point):
Did the introduction clearly identify the core issue(s) in the problem, and was the overall logical structure systematic?&A\\
10&Analysis of Facts in Relation to Legal Basis (1 point):
Did the answer analyze the facts presented in the problem (e.g., plaintiff’s claims) in light of the relevant legal provisions, and was the conclusion logically derived?&C\\
\bottomrule
\end{tabularx}
\vspace{1em}
\end{table*}
\endgroup
\begin{figure*}[t]
    \centering

    \begin{subfigure}{\textwidth}
        \centering
        \includegraphics[width=\textwidth]{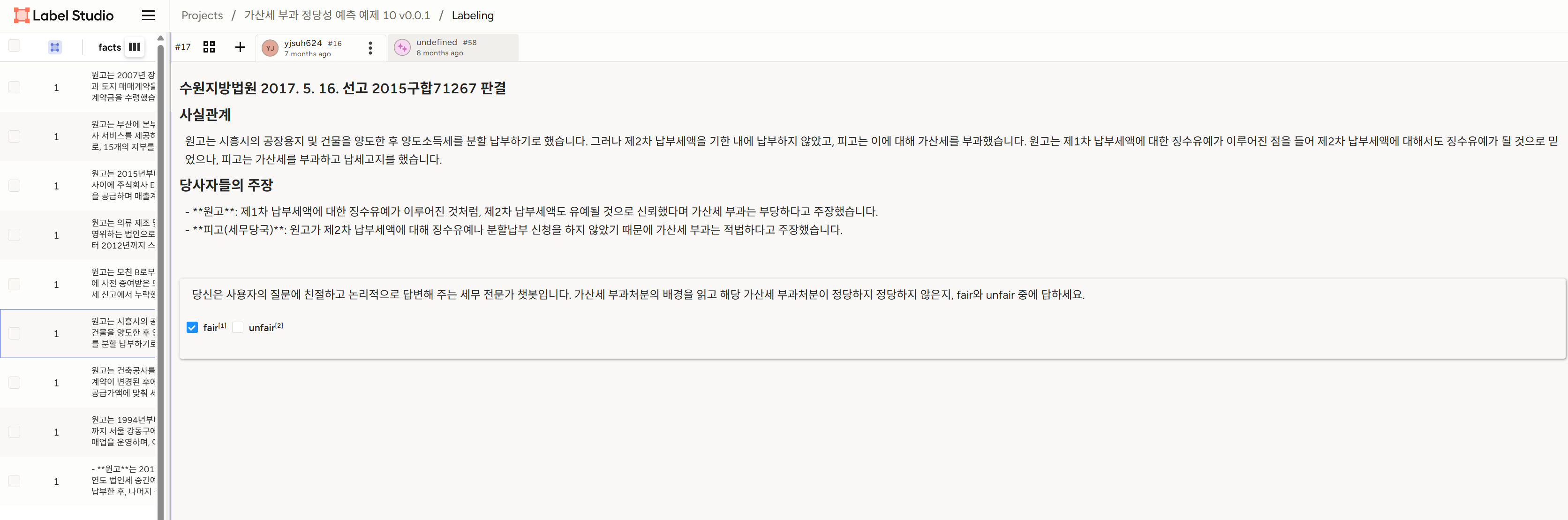}
        \caption{Label Studio interface for \oursthree.}
        \label{fig:labelstudio}
    \end{subfigure}

    \vspace{1em} 

    \begin{subfigure}{\textwidth}
        \centering
        \includegraphics[width=\textwidth]{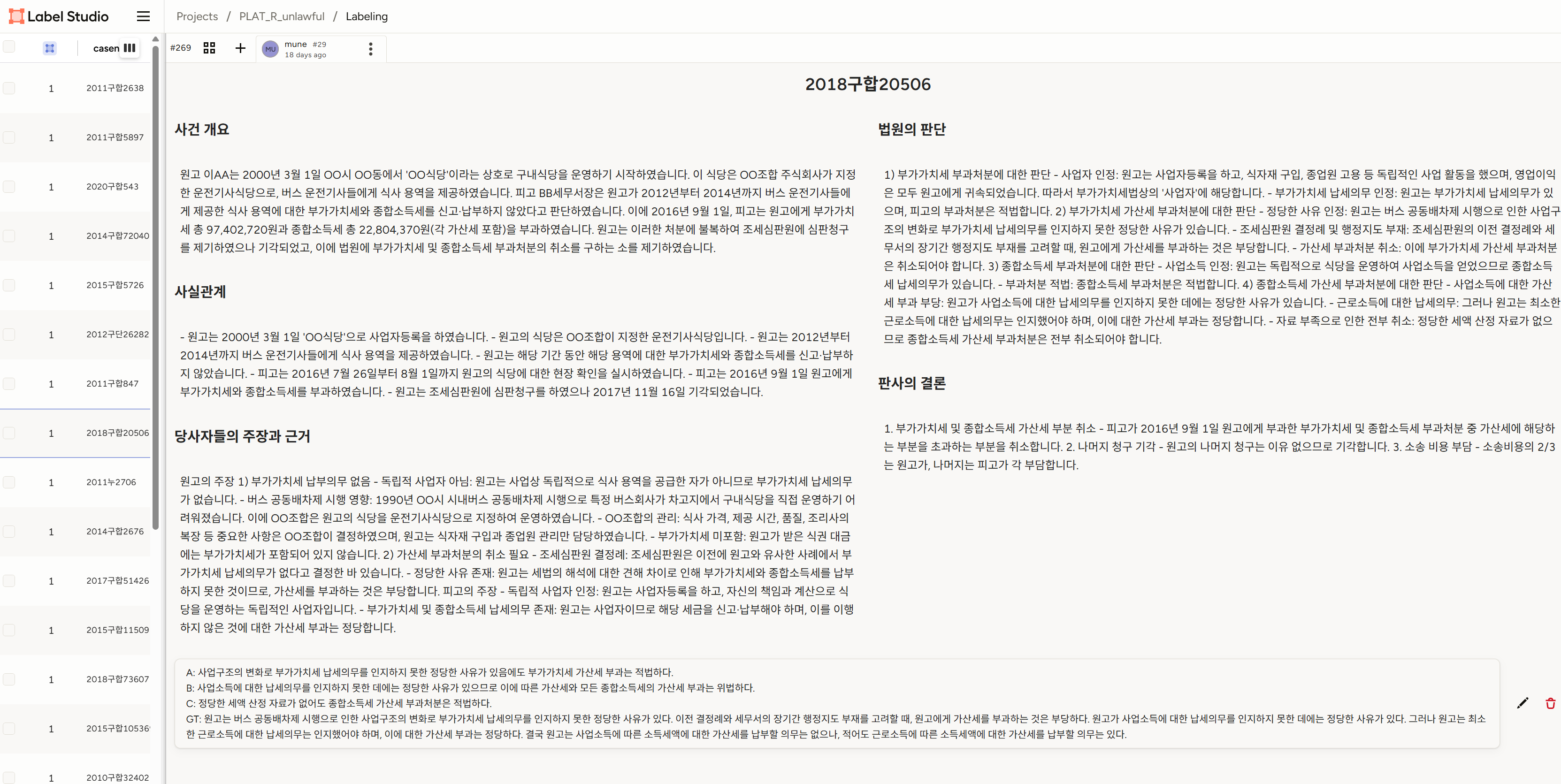}
        \caption{Label Studio interface for \oursfour.}
        \label{fig:labelstudio2}
    \end{subfigure}

    \caption{Label Studio interfaces for our datasets. Each subfigure shows annotation setup for (\subref{fig:labelstudio}) \oursthree{} and (\subref{fig:labelstudio2}) \oursfour{}.}
    \label{fig:labelstudio_combined}
\end{figure*}

\begingroup
\begin{figure*}[t]
\centering
\begin{tcolorbox}[promptstyle]
{\small 
\smallskip
\textbf{System Prompt}\par
\smallskip
    You are a tax expert chatbot that provides friendly and logical answers to users' questions.\\
\smallskip

\textbf{Vanilla User Prompt}\par
\smallskip

    Please read the background and materials related to the imposed penalty tax presented above. 
    
    Determine whether the penalty tax is “Legitimate”, “Not legitimate”, or, if a clear conclusion cannot be reached, answer “don’t know". Then, provide an explanation for your choice.

    \textbf{Problem description}: \{precedent\}\\
\smallskip

\textbf{LLM with RAG User Prompt}\par
\smallskip

    Please read the background and materials related to the imposed penalty tax presented above. 
    
    Determine whether the penalty tax is “Legitimate”, “Not legitimate”, or, if a clear conclusion cannot be reached, answer “Don't know”. Then, provide an explanation for your choice.

    \textbf{Problem description}: \{precedent\}   
    
    \textbf{Reference material}: \{raged doc\} 
    
    \textbf{\#\#\#Answer:}
} 
\end{tcolorbox}
\caption{Prompts to Answer \oursthree{}: Vanilla vs. with RAG}
\label{promptmc}
\end{figure*}

\begin{figure*}[t]
\centering
\begin{tcolorbox}[promptstyle]
{\small 
\textbf{System Prompt}\par
\smallskip
    You are a tax expert chatbot that provides friendly and logical answers to users' questions.\\

\textbf{User Prompt}\par
\smallskip
The following is a multiple-choice question on the imposition of additional tax.
Read the case and each option carefully, then choose only one option (A, B, C, or D)
that is most correct regarding the imposition of the additional tax.
You must respond in the format \texttt{Answer: A} and provide an explanation.\\
\textbf{Question}: \{question\}\\
\textbf{A}. \{option\_A\}\\
\textbf{B}. \{option\_B\}\\
\textbf{C}. \{option\_C\}\\
\textbf{D}. \{option\_D\}\\
\textbf{\#\#\# Answer:}
} 
\end{tcolorbox}
\caption{Prompt to Answer \oursfour.}
\label{promptplatmc4}
\end{figure*}

\begin{figure*}[t]
\centering
\begin{tcolorbox}[promptstyle]
{\small 
\textbf{System Prompt}\par
\smallskip
    You are a tax expert chatbot that provides friendly and logical answers to users' questions.\\

\textbf{User Prompt}\par
\smallskip
You are an expert in tax law and must address legal issues in a systematic, exam-style manner.
Unless otherwise specified, assume that South Korean tax law applies. However, where contextually justified, you should also address legal issues beyond South Korean tax law.
Use precise legal terminology, and always use honorifics in your responses.
Do not mention disclaimers or the need for external legal advice.
Do not ask the user to look up the law themselves.
Provide focused legal analysis and individualized advice.
Your answers should be decisive and authoritative; do not state that it is merely general information.
You must include Korea-specific legal terminology.
If you identify relevant legal considerations, present a concise and clear legal analysis.
You must always cite specific statutory provisions, specifying Article (조), Paragraph (항), Subparagraph (호), and Item (목).
Example: “Article 6, Paragraph 1 of the Framework Act on National Taxes.”
Using only general references (e.g., “Framework Act on National Taxes”) is not permitted.
If there is no relevant information, you must explicitly state that none exists.
If reliable sources are available, share substantive guidance or insights.
The response must be in the same language as the question.
If the question requests a “simple answer,” you must provide a concise answer.
If you are asked to analyze a specific case from the certified tax accountant examination, but the case text or details are not provided, you must explicitly point out that the essential case material is missing.\\
\\
\textbf{Read the following case and answer the question.  
Explain whether the imposition of additional tax (penalty tax) in the above case is lawful.}  \smallskip

\textbf{(*Note: Only the lawfulness of the imposition of additional tax is to be considered; no other issues are to be taken into account.*) } 
\smallskip

\textbf{<Case>} \{precedent\}
} 
\end{tcolorbox}
\caption{Prompt to Answer \oursessay.}
\label{promptessay}
\end{figure*}

\begin{figure*}[t]
\centering
\begin{tcolorbox}[promptstyle]
{\small 
\textbf{User Prompt}\par
\smallskip

    You are a grader objectively and officially evaluating answers for a Korean Tax Law exam. You are an expert in tax law, and you must handle legal issues systematically and in an exam style.

\smallskip
    
    \textbf{Goal}: Your task is to determine how closely the candidate’s answer aligns with the model answer provided. Refer to the model answer to evaluate accuracy, completeness, and legal reasoning.
\smallskip

    \textbf{Context}: You will be given both a candidate’s answer and a model answer.
\smallskip

    \textbf{Output Format}:
    
    After reviewing the answer:
    
    Explanation: Briefly explain how the candidate’s answer aligns with or deviates from the model answer.
    
    Scoring: The score must reflect how well the candidate’s answer fulfills the requirements of the model answer, using the following strict format:
    
    Example: “Comprehensive Score: [[5]]”
    
    Comprehensive Score: Assign a final score between 0 and 10 (whole numbers only).
    
    - 10 = fully satisfactory (100\%)
    
    - Lower scores proportionally reflect deficiencies (e.g., 5 = 50\% fulfilled).
    
    - If the candidate includes material not in the model answer, you must generally deduct points unless you are certain it is legally accurate and relevant.
    
    - Assume the model answer contains all information necessary for a complete response.
    
    - If the model answer cites books or academic articles, candidates do not need to reproduce them.
    
    - However, statutory provisions must be cited accurately with precise identification of Article, Paragraph, Subparagraph, Item, etc.
    \smallskip

    \textbf{Reference}: \{Grading Rubrics\}
} 
\end{tcolorbox}
\caption{Prompt for LLM-as-a-Judge Scoring}
  \label{llm_judge_prompt}
\end{figure*}

\begin{figure*}[t]
\centering
\begin{tcolorbox}[promptstyle]
{\small 

    \textbf{System Prompt}
    
    \smallskip
    Here are the rules you must always follow to complete the task successfully:
    
    You must provide at least one tool invocation. If you do not, the task will fail.
    
    Use the correct arguments for each tool. Do not pass variable names as arguments--always use actual values.
    
    Only call tools when necessary. If you already have enough information, do not call the search agent. Try to solve the task yourself first.
    
    The more tools you call, the more hints you will gather, which will guide you toward the correct final answer.

    You must call at least two different tools besides the final answer tool. If you can determine the final answer, return it using the final answer tool.
    
    The retrievertool must be called at least three times. The retrievertool works best in synergy with other tools. So, whenever you call retrievertool, follow up with calls to other relevant tools.
    
    Do not repeat tool calls with the exact same parameters as a previous invocation.

\smallskip
\smallskip
    
    \textbf{User Prompt}
    
    \smallskip
    Please read the background and materials related to the imposed penalty tax presented above. Based on this information, determine whether the penalty tax is “Legitimate”, “Not legitimate”, or, if a clear conclusion cannot be reached, answer “Unknown”. Then, provide an explanation for your choice.
\smallskip

    \textbf{Problem description}: \{precedent\}
\smallskip

    \textbf{Reference material}: \{retrieved doc\}
\smallskip

    \textbf{\#\#\# Answer:}
} 
\end{tcolorbox}
\caption{Prompt for Agentic RAG (The default prompt from ToolCallAgent of \texttt{smolagents} library is used.)}
  \label{promptar}
\end{figure*}

\begin{figure*}[t]
\centering
\begin{tcolorbox}[promptstyle]
{\small 

\textbf{User Prompt}\par
\smallskip
    The above document contains statutes and precedents retrieved in relation to penalty taxes.  
    You must carefully review the document. 
    \smallskip
    
    \textbf{Summarize the relevant statutes and precedents according to the following format}: \\
    \smallskip
    Format: \\
    The relevant statutes are as follows. “Statute1 ...” \\
    The parts of the statute that are relevant to the issue are as follows. “Relevance1-1 ... Relevance1-2 ...” \\
    The parts of the statute that are not relevant to the issue are as follows. “Irrelevance1-1 ... Irrelevance1-2 ...” \\
    
    The relevant precedents are as follows. “Precedent1 ...” \\
    The parts of the precedent that are relevant to the issue are as follows. “Relevance2-1 ... Relevance2-2 ...” \\
    The parts of the precedent that are not relevant to the issue are as follows. “Irrelevance2-1 ... Irrelevance2-2 ...” \\

    Generate only the results you identified from the document. 
    Do not include any additional explanations. \\

} 
\end{tcolorbox}
\caption{Prompt for Document-Relevance-Analyzer tool}
\label{promptmadc}
\end{figure*}

\begin{figure*}[t]
\centering
\begin{tcolorbox}[promptstyle]
{\small 

\textbf{User Prompt}\par
\smallskip
You are a skilled legal expert tasked with evaluating legal reasoning responses.
Use the given context to answer the question accurately and naturally.
You must strictly adhere to the following formatting rules:
\smallskip
After completing your analysis and reasoning, the final line of your response must be in the format:
“The answer is {final conclusion}”

Do not include any additional explanation or reasoning after the phrase “The answer is”.

The phrase “The answer is” must appear exactly once, and only as the last line of your response.\\
\smallskip

\textbf{Analyze the given legal case scenario by following the structured steps below:\\
}
<issue> Identify the key legal issue in the case. </issue>

<rule> Clearly state the statutes or legal principles that govern the identified issue. </rule>

<application> Analyze how the above rules or principles apply to the specific facts of the case. Discuss the legal validity of the case based on that application. </application>

<conclusion> Synthesize your analysis and provide the likely legal conclusion based on the application of law to the issue. </conclusion>

} 
\end{tcolorbox}
\caption{Prompt for Legal-Analyzer tool}
\label{promptmala}
\end{figure*}

\begin{figure*}[t]
\centering
\begin{tcolorbox}[promptstyle]
{\small 
\textbf{User Prompt (Plaintiff's lawyer role-playing tool)}\par
\smallskip

    You are a lawyer representing the plaintiff (the taxpayer) in a tax case. For the following issue, argue unconditionally from the taxpayer's perspective that the imposition of the penalty tax is not legitimate.\\
\smallskip

\textbf{User Prompt (Defendant's lawyer role in role-playing tool)}\par
\smallskip

    You are a lawyer representing the defendant (the tax authority) in a tax case. For the following issue, argue unconditionally from the tax authority's perspective that the imposition of the penalty tax is legitimate.\\
\smallskip

\textbf{User Prompt (Judge role in role-playing tool)}\par
\smallskip

    You are a neutral tax judge. Below are the arguments from both parties:
    
    [Plaintiff's Argument]
    {claim-a}
    
    [Defendant's Argument]
    {claim-b}
    
    Please compare the arguments from both sides, evaluate their validity, and reach a final conclusion based on legality and logical reasoning. If you identify any flaws, point them out and correct them to present the proper conclusion.\\

} 
\end{tcolorbox}
\caption{Prompt for Virtual-court tool}
\label{promptvc}
\end{figure*}

\begin{figure*}[t]
\centering
\begin{tcolorbox}[promptstyle]
{\small 
\textbf{[Query]}\par
\smallskip

    Please read the background and materials related to the imposed penalty tax presented above. Determine whether the penalty tax is “Legitimate”, “Not legitimate”, or, if a clear conclusion cannot be reached, answer “don’t know". Then, provide an explanation for your choice. Problem description: \{precedent\}\\
\smallskip

\textbf{[Iterative retrieval steps]}\par
\smallskip

Observations: 

===== Document 0 =====

Inheritance Tax and Gift Tax Act Article 15 (Provisional inclusion of disposed property, etc., before commencement of inheritance) ① Where the decedent has disposed of property or incurred liabilities, and any of the following subparagraphs applies, such property or amount shall be presumed to have been inherited and shall be included in the taxable value of inheritance tax under Article 13. 1. Where the amount received by the decedent from disposing of property, or the amount withdrawn from the decedent’s property, calculated by each type of property, is 200 million won or more within 1 year before the commencement date of inheritance, …

===== Document 1 =====

This is a precedent in which the imposition of penalty tax is lawful. Legal basis: Framework Act on National Taxes Article 14(1) (principle of taxation based on substance) (2) Error in calculating the tax base and applying the tax rate

Claim: The sales proceeds actually received by the plaintiff are only about 232.2 million won. …

===== Document 2 =====

This is a precedent in which the imposition of penalty tax is lawful. According to the proviso of Article 49(1) of the Enforcement Decree of the Inheritance Tax and Gift Tax Act, …
\\
\smallskip

\textbf{[Document-Relevance-Analyzer tool’s intermediate outputs]}\par
\smallskip

Observations: 

==========| Document Inspection Start ]==========
‘The relevant statutes are as follows.   ”Inheritance Tax and Gift Tax Act Article 15 Paragraph 1” The parts of the above statute that are related to the issue are as follows.
 ”Related Point 1-1: A 200 million won deduction rule is explicitly stated, but direct documents on actual …

”Unrelated Point 1-1: Clear provisions on other deductions besides the 200 million won deduction (minimum cost of living deduction …

==========| Document Inspection End ]============\\
\smallskip

\textbf{[Legal-Analyzer tool’s intermediate outputs]}\par
\smallskip

Observations: 

==========| Legalanalyzer Inspection Start ]==========

<issue>

Whether the 200 million won deduction provision of Inheritance Tax and Gift Tax Act Article 15 Paragraph 1 applies to this case, … 

</issue>

<rule>

Inheritance Tax and Gift Tax Act Article 15 Paragraph 1 provides that 200 million won shall be deducted from the taxable value of inheritance tax. This corresponds to … 

</rule>

<application>

First, the 200 million won deduction provision of Inheritance Tax and Gift Tax Act Article 15 Paragraph 1 is a basic deduction applied when calculating the taxable value of inheritance tax ...

</application>

<conclusion>

Therefore, the 200 million won deduction provision of Inheritance Tax and Gift Tax Act Article 15 Paragraph 1 applies to the calculation of the taxable value of inheritance tax, and the claims … 

</conclusion>

==========| Legalanalyzer Inspection End ]============\\
\smallskip

\textbf{[Final reasoning]}\par
\smallskip

Final answer: “unlawful”

[Explanation and grounds]
\begin{itemize}
    \item The 200 million won deduction provision in Article 15(1) of the Inheritance Tax and Gift Tax Act corresponds to a basic deduction applied when calculating the taxable value of inheritance tax, and it may also apply to this case. However, whether this deduction applies does not directly affect the lawfulness of imposing penalty tax (especially the failure-to-pay penalty tax).
    \item The minimum cost of living deduction and the debt repayment amount deduction may likewise be recognized if their separate requirements are satisfied when calculating the taxable value of inheritance tax, but these deduction claims do not directly affect the lawfulness of imposing penalty tax.
    \item …
\end{itemize}

} 
\end{tcolorbox}
\caption{Worked example in Agentic Retrieval}
\label{workedexampleinagenticretrieval}
\end{figure*}

\begin{table*}[!htbp]
\centering
\scriptsize
\caption{Worked examples of Models' common failure patterns 1}
\label{workedexfailurepatterns-case1}
    \vspace{-2mm}
\begin{tabularx}{\textwidth}{l X}
\toprule
\textbf{Case 1. 2016nu11 (Lawfulness: True)} \\
\midrule
\textbf{Facts} & 1. Plaintiff AAA was an executive of JJJ Corporation. 2. On December 29, 2003, Plaintiff AAA received 00,000 shares of JJJ, an unlisted company, as a gift from the largest shareholder LLL. 3. Plaintiff AAA participated in paid-in capital increases a total of three times on October 5, 2004, November 5, 2004, and December 22, 2005, acquiring a total of 00,000 shares at 000 won per share. 4. Plaintiffs EEE, FFF, BBB, CCC, GGG, HHH, and DDD were employees of JJJ. 5. On December 1, 2004, they acquired 0,000 shares of JJJ each for consideration from the largest shareholder MMM. 6. They participated in the paid-in capital increase on December 22, 2005, acquiring 000 shares each at 0,000 won per share. 7. JJJ split the par value per share at 000 won on July 1, 2007, and was listed on the KOSDAQ market on January 25, 2008. 8. The number of shares held by the plaintiffs increased tenfold due to the stock split. 9. The ○○ Regional National Tax Service conducted a comprehensive tax audit of JJJ from August 10 to September 30, 2009. 10. After the investigation, gift tax was initially imposed on the plaintiffs for the profit from the listing of the shares acquired, but no gift tax was imposed on the profit from the new shares issued for consideration. 11. On January 25, 2010, the Board of Audit and Inspection requested the ○○ Regional National Tax Service to take corrective action by taxing the listing profit of the new shares issued for consideration. 12. The defendant included the listing profit of the new shares issued for consideration in the value of the gifted property and imposed gift tax and additional tax on the plaintiffs (hereinafter referred to as "the disposition in this case"). 13. The plaintiffs filed an appeal with the Tax Tribunal, contesting the disposition in this case, but it was dismissed. \\
\\[1ex]
\textbf{Claims} & \textbf{Plaintiffs' Arguments} 1. Claim of Justifiable Cause: - Whether to impose gift tax on listing gains from stock issued through paid-in capital increase is a matter of conflicting opinions even among tax authorities. - The defendant decided not to impose gift tax on new shares issued through paid-in capital increase during the tax audit on August 10, 2009. - The imposition of penalties is illegal because there is a justifiable reason for the plaintiffs' failure to fulfill their reporting and payment obligations. 2. Claim of Trust in Tax Officials: - The plaintiffs did not pay gift tax on the listing gains from new shares issued through paid-in capital increase due to the defendant's erroneous tax audit result notification. - This is the defendant's fault, so the imposition of the failure-to-pay penalty is unfair. 

\textbf{Defendant's Arguments} 1. Claim of Legality of Penalty Imposition: - There is no justifiable reason for the plaintiffs. - Under tax law, penalties are administrative sanctions imposed regardless of the taxpayer's intent or negligence. - Therefore, the imposition of penalties is lawful. \\
\\[1ex]
\textbf{Reasoning} & 1. Legal Principles of Imposition of Additional Tax: - Additional tax is imposed according to Article 47-2 (Tax for Non-filing), Article 47-3 (Tax for Under-reporting), and Article 47-5 (Tax for Failure to Pay) of the former Framework Act on National Taxes. - According to Article 48, Paragraph 1 of the same Act, additional tax shall not be imposed if there is a justifiable reason for non-compliance with the tax obligation. 2. Non-existence of Justifiable Reason: - Additional tax under tax law is an administrative sanction imposed when a taxpayer violates the obligation to report and pay without a justifiable reason, and the taxpayer's intent or negligence is not considered. - Ignorance or misinterpretation of laws and regulations does not constitute a justifiable reason. \textbf{- Even if the obligation to report and pay is not fulfilled due to reliance on incorrect explanations from tax officials, it does not constitute a justifiable reason if it is clear that it contradicts related laws and regulations.} 3. Taxability of Paid-in New Shares: - It is also specified in the "Tax Audit Casebook on Stock Transactions" published by the National Tax Service in March 2008 that the listing profit of shares from paid-in capital increase is subject to gift tax. - In light of the wording and legislative intent of Article 41-3, Paragraph 1 and Paragraph 6 of the former Inheritance and Gift Tax Act, it is clear that the listing profit of paid-in new shares is subject to gift tax. 4. Conclusion: - The justifiable reason claimed by the plaintiffs cannot be recognized.

\textbf{Therefore, the imposition of additional tax is lawful.}\\[1ex]
\midrule
\textbf{Model's decision}\\
Claude3.7-sonnet & \textbf{Not Lawful.}

Existence of Justifiable Cause: The plaintiffs argue that a justifiable cause exists because whether to levy gift tax on the listing gains of stocks acquired through a paid-in capital increase was a matter of conflicting views even among tax authorities, and the defendant decided not to impose gift tax on the new shares during the initial tax audit. However, the defendant argues that penalty tax is an administrative sanction imposed regardless of the taxpayer's intent or negligence.

\textbf{Protection of Reliance on Tax Officials}: The plaintiffs argue that they did not pay the gift tax on the listing gains of the new shares due to the defendant’s incorrect notification of tax audit results. This is an assertion of reliance on the provision of incorrect information by tax officials.

Interpretation and Application of Tax Law: In principle, penalty taxes are generally imposed regardless of the taxpayer's intent or negligence. However, if a taxpayer fails to fulfill the obligation of filing and payment by relying on incorrect information or guidelines from the tax authorities, justifiable cause may be recognized. In this case, the plaintiffs acted according to the tax authority's initial judgment, and since this was due to the tax authority's mistake, there is room to see that a justifiable cause exists for the plaintiffs. However, the defendant is arguing that, under tax law, penalty tax is a sanction imposed regardless of intent or negligence.

Conclusion: "Not Lawful"

Since the plaintiffs acted according to the initial judgment of the tax authorities, and this was due to the tax authorities' mistake, it can be seen that there is a justifiable cause for the plaintiffs. Therefore, the imposition of penalty taxes is highly likely to be not lawful.
\\
GPT-4o & \textbf{Not Lawful.}

Existence of Justifiable Cause

The issue of whether to levy gift tax on listing gains from shares acquired through a paid-in capital increase was a complex matter where opinions conflicted even within the tax authority.

During the tax audit process, the tax authority itself decided not to impose gift tax on the new shares, but the taxation policy was later changed following a correction request from the Board of Audit and Inspection.

Application of the Principle of Protection of Legitimate Expectations

The taxpayers acted in reliance on the tax authority’s original tax audit results.

For the tax authority to reverse its own decision and cause disadvantage to the taxpayers \textbf{violates the principle of protection of legitimate expectations.}

Grounds for Exception to the Imposition of Penalty Taxes

The Supreme Court takes the position that "penalty taxes cannot be imposed if the taxpayer has a justifiable cause other than ignorance or misunderstanding of the law."

In this case, a justifiable cause exists in the form of a change in the tax authority's authoritative interpretation.

Therefore, it is judged that imposing penalty taxes on the plaintiffs is not lawful.
\\
\bottomrule
\end{tabularx}
\end{table*}
\vspace{1em}

\begin{table*}[!htbp]
\centering
\scriptsize
\caption{Worked examples of Models' common failure patterns 2}
\label{workedexfailurepatterns-case2}
    \vspace{-2mm}
\begin{tabularx}{\textwidth}{l X}
\toprule
\textbf{Case 2. 2018guhap20506 (Lawfulness: False)}\\
\midrule
\textbf{Facts} & The Plaintiff, Lee AA, began operating a cafeteria under the trade name 'OO Restaurant' in OO-dong, OO-city, on March 1, 2000. This restaurant was a driver-only cafeteria designated by the OO Association Co., Ltd. and provided meal services to bus drivers.

The Defendant, the Head of the BB Tax Office, determined that the Plaintiff had failed to report and pay Value Added Tax (VAT) and Comprehensive Income Tax on the meal services provided to bus drivers from 2012 to 2014. Consequently, on September 1, 2016, the Defendant assessed the Plaintiff a total of 97,402,720 KRW in VAT and 22,804,370 KRW in Comprehensive Income Tax (both including penalty taxes).

The Plaintiff filed an appeal with the Tax Tribunal contesting these dispositions, but the petition was dismissed. Subsequently, the Plaintiff filed a lawsuit seeking the cancellation of the VAT and Comprehensive Income Tax assessments. 1. The Plaintiff registered as a business under the name 'OO Restaurant' on March 1, 2000. 2. The Plaintiff's restaurant is a designated driver-only cafeteria for the OO Association. 3. The Plaintiff provided meal services to bus drivers from 2012 to 2014. 4. During said period, the Plaintiff did not report or pay VAT and Comprehensive Income Tax for those services. 5. The Defendant conducted a field inspection of the Plaintiff's restaurant from July 26 to August 1, 2016. 6. On September 1, 2016, the Defendant imposed VAT and Comprehensive Income Tax on the Plaintiff. 7. The Plaintiff filed an appeal with the Tax Tribunal, which was dismissed on November 16, 2017. \\
\\[1ex]

\textbf{Claims} & \textbf{Plaintiffs' Arguments} 1. No VAT payment obligation - Not an independent business operator: The plaintiff is not subject to VAT as they are not a person who independently supplied meal services. - Impact of the bus joint dispatch system: It became difficult for specific bus companies to directly operate in-house restaurants at the garage due to the implementation of the OO City intra-city bus joint dispatch system in 1990. Accordingly, the OO Association designated the plaintiff's restaurant as a driver's restaurant and operated it. - Management by the OO Association: The OO Association determined important matters such as meal prices, serving hours, quality, and chef's attire, and the plaintiff was only in charge of purchasing food ingredients and managing employees. - VAT not included: VAT is not included in the meal ticket payments received by the plaintiff. 2. Need to cancel the penalty surcharges - Tax Tribunal ruling: The Tax Tribunal has previously ruled that there is no VAT payment obligation in cases similar to the plaintiff's. - Existence of justifiable reasons: It is unfair to impose penalty surcharges because the plaintiff failed to pay VAT and comprehensive income tax due to differences in interpretation of tax laws.

\textbf{Defendant's Arguments} 1. Claim of Legality of Penalty Imposition: - There is no justifiable reason for the plaintiffs. - Under tax law, penalties are administrative sanctions imposed regardless of the taxpayer's intent or negligence. - Therefore, the imposition of penalties is lawful. \\
\\[1ex]
\textbf{Reasoning} & ... Judgment on Value-Added Tax Penalty Assessment \textbf{- Acknowledgment of Justifiable Cause: The plaintiff had a justifiable reason for not recognizing the obligation to pay value-added tax due to changes in the business structure resulting from the implementation of the bus joint dispatch system. - Precedent of the Tax Tribunal and Absence of Administrative Guidance: Considering the previous decisions of the Tax Tribunal and the long-term absence of administrative guidance from the tax office, it is unreasonable to impose a penalty on the plaintiff. - Cancellation of Penalty Assessment: Therefore, the value-added tax penalty assessment should be canceled.} ... \textbf{Judgment on Comprehensive Income Tax Penalty Assessment - Unjustified Penalty on Business Income: There is a justifiable reason for the plaintiff not recognizing the obligation to pay taxes on business income. - Obligation to Pay Taxes on Labor Income: However, the plaintiff should have at least recognized the obligation to pay taxes on labor income, and the penalty assessment for this is justified. - Full Cancellation Due to Lack of Data: Since there is no data to calculate the correct tax amount, the comprehensive income tax penalty assessment must be canceled in full.}
\\[1ex]
\midrule
\textbf{Model's decision}\\
Claude3.7-sonnet & \textbf{Lawful.}

\textbf{Recognition of Independent Business Operator Status}: The plaintiff has registered the business under their own name and operated the restaurant since 2000. The plaintiff directly performed the purchase of food ingredients and employee management, and managed revenues and expenses under their own account. \textbf{Facts of Failure to Fulfill Tax Obligations}: The plaintiff failed to report and pay Value Added Tax and Comprehensive Income Tax from 2012 to 2014. It is clear that the basic tax obligations as a business operator were not fulfilled. Inappropriateness of the Plaintiff’s Arguments: The fact that there was management and supervision by the OO Association is difficult to serve as a ground for denying independent business operator status. Although it is claimed that there are Tax Tribunal decisions in past similar cases, each case must be judged individually. The argument regarding the difference in the interpretation of tax laws is not sufficient to justify the failure to fulfill basic tax obligations. \textbf{Legitimacy of the Penalty Tax Imposition}: \textbf{The imposition of penalty taxes as a sanction for the violation of tax obligations is a natural consequence under tax law. No special circumstances that could be recognized as a "justifiable cause" were found.} \textbf{Therefore, the disposition to impose the relevant penalty taxes is judged to be lawful.}
\\
GPT-4o & \textbf{Lawful.}

To determine whether the imposition of penalty taxes in this case is lawful, the focus must be on whether the plaintiff is an independent business operator and whether there is an obligation to pay taxes.

Whether an Independent Business Operator: The defendant argues that the plaintiff is an independent business operator who registered the business and operated the restaurant under their own responsibility and account. On the other hand, the plaintiff argues that they are not an independent business operator because the OO Association determined the major matters of the restaurant's operation. \textbf{However, the fact that the plaintiff completed the business registration and directly handled the purchase of food ingredients and employee management can serve as grounds to view them as an independent business operator.}

Tax Payment Obligation: The plaintiff did not report or pay Value Added Tax and Comprehensive Income Tax. The defendant argues that since the plaintiff is a business operator, they have an obligation to report and pay the corresponding taxes. \textbf{Generally, when a business registration is made, the obligation to pay Value Added Tax and Comprehensive Income Tax arises.}

Justification for the Imposition of Penalty Taxes: The plaintiff argues that the imposition of penalty taxes is unjust, claiming they failed to pay taxes due to differences in the interpretation of tax laws. However, it is difficult to recognize the imposition of penalty taxes as unjust based solely on differences in the interpretation of tax laws. Especially if it is not clear whether the Tax Tribunal has previously made a different decision regarding a similar case, the imposition of penalty taxes may be justified.

\textbf{In conclusion, considering that the plaintiff registered the business and operated the restaurant, and failed to fulfill the obligation to report and pay taxes, the imposition of penalty taxes is judged to be lawful.}
\\
\bottomrule
\end{tabularx}
\end{table*}
\endgroup

\begin{figure*}[t]
    \centering
    \includegraphics[width=16cm]{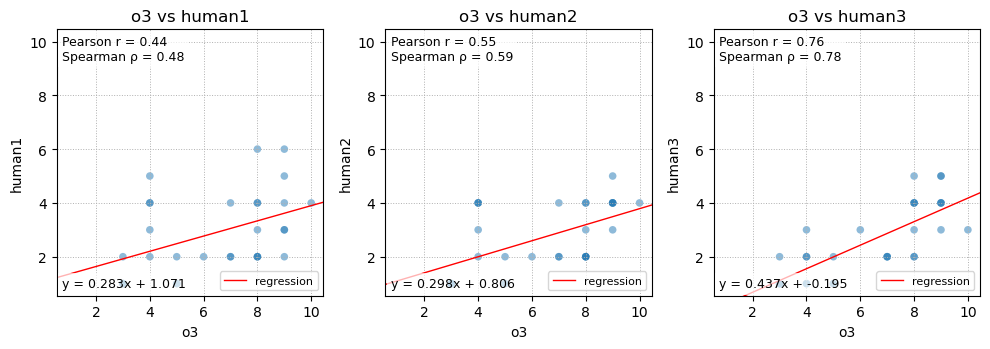}
    \caption{Scatter plots comparing the scores assigned by the LLM-based grader (o3) with those of three human graders. Each panel shows the regression line (red) with the corresponding Pearson’s r, Spearman’s ρ, regression equation and . While o3 scores are positively correlated with all human graders, the strength of correlation varies across graders (r = 0.44, 0.55, and 0.76, respectively).}
    \label{red_scatter3}
\end{figure*}


\renewcommand{\arraystretch}{1.15}
\setlength{\tabcolsep}{6pt}

\begin{table*}[htbp!]
\centering
\resizebox{\textwidth}{!}{%
\begin{tabular}{@{} l c l @{}}
\toprule
\textbf{Name} & \textbf{License} & \textbf{URL} \\
\midrule
EXAONE3.5-32B \cite{research2024exaone35serieslarge} & MIT License & \url{https://huggingface.co/LGAI-EXAONE/EXAONE-3.5-32B-Instruct} \\
Qwen3-32B \cite{yang2025qwen3technicalreport} & CC BY 4.0 & \url{https://huggingface.co/Qwen/Qwen3-32B} \\
smolagents \cite{smolagents} & MIT License & \url{https://github.com/huggingface/smolagents} \\
\textbf{\ours (ours)} & CC BY-NC & Dataset will be released on GitHub upon publication. \\
\bottomrule
\end{tabular}}
\caption{Datasets and models used in this work, with licenses.}
\label{license}
\end{table*}

\end{document}